\documentclass[12pt]{article}

\usepackage{hyperref}
\usepackage{verbatim}

\newcommand{\Z}{\mathbb{Z}}

\newcommand{\calG}{\mathcal{G}}
\newcommand{\calS}{\mathcal{S}}
\newcommand{\calW}{\mathcal{W}}

\newcommand{\Bbar}{\bar{B}}

\newcommand{\Khat}{\widehat{K}}
\newcommand{\dhat}{\widehat{d}}
\newcommand{\Xhat}{\widehat{X}}
\newcommand{\thetahat}{\widehat{\theta}}

\newcommand{\rhohat}{\widehat{\rho}}
\newcommand{\Chat}{\widehat{\mathcal{C}}_{KC}}
\newcommand{\CKC}{\mathcal{C}_{KC}}
\newcommand{\synthMB}{\texttt{synthMB}}

\newcommand{\synthSemiparMB}{\texttt{synthSemiparMB}}

\newcommand{\Phat}{\widehat{P}}

\newcommand{\pisim}{\sim_\pi}
\newcommand{\motifsim}{\sim_M}

\newcommand{\mclustase}{GMM \circ ASE}
\newcommand{\smclustase}{SemiparGMM \circ ASE}

\usepackage{graphics}
\usepackage{graphicx}
\usepackage{bbm}
\usepackage{caption}
\usepackage{subcaption}
\usepackage[round,authoryear,comma]{natbib}
\usepackage{parskip}
\usepackage{mathrsfs}
\usepackage{fullpage}
\usepackage{amsfonts,latexsym,eucal,amsmath,amsthm,amssymb,bm,pdfpages}
\usepackage{array}
\usepackage[T1]{fontenc}
\usepackage[ansinew]{inputenc}
\usepackage{enumerate}
\usepackage{booktabs}
\usepackage{algorithm,algorithmic}
\usepackage{rotfloat}
\usepackage{booktabs}
\usepackage{multirow}
\usepackage{listings}

\renewcommand{\hat}{\widehat}

\renewcommand{\Re}{\mathbb{R}}

\newcommand{\p}{\mathbb{P}}

\newcommand{\mC}{\mathcal{C}}

\newtheorem{theorem}{Theorem}

\newtheorem{corollary}[theorem]{Corollary}
\theoremstyle{definition}
\newtheorem*{remark*}{Remark}
\newtheorem{definition}[theorem]{Definition}

\renewcommand{\Re}{\mathbb{R}}

\graphicspath{{./}{}}%
\graphicspath{{./Figures/}{}}%
\graphicspath{{./INIslides/}{}}%

\newcommand{\blind}{1}

\begin{document}

\def\spacingset#1{\renewcommand{\baselinestretch}%
{#1}\small\normalsize} \spacingset{1}

\if1\blind
{
\title{
%\bf Semiparametric spectral modeling \\ of the \\ {\it Drosophila} connectome
\bf Semiparametric spectral modeling \\ of the {\it Drosophila} connectome
}

\author{
%Carey E. Priebe\\Youngser Park\\Minh Tang\\Avanti Athreya\\Vince Lyzinski\\Joshua Vogelstein\\
%Carey E. Priebe, Youngser Park, Minh Tang, Avanti Athreya, Vince Lyzinski, Joshua Vogelstein\\ 
C.E. Priebe, Y. Park, M. Tang, A. Athreya, V. Lyzinski, J.T. Vogelstein\\ 
%Department of Applied Mathematics and Statistics\\
Johns Hopkins University\\
%\&\\
 \\
Yichen Qin\\
University of Cincinnati\\
%\&\\
 \\
Ben Cocanougher\\
HHMI Janelia Research Campus \& Cambridge University, UK\\
 \\
Katharina Eichler\\
HHMI Janelia Research Campus \& University of Konstanz, Germany\\
%Baltimore, Maryland, USA\\
%\&\\
 \\
%Marta Zlatic\\Albert Cardona\\
Marta Zlatic and Albert Cardona\\
HHMI Janelia Research Campus\\
%Baltimore, Maryland, USA\\
}
\maketitle
} \fi

\if0\blind
{
  \bigskip
  \bigskip
  \bigskip
  %\begin{center}
    %{\LARGE\bf Semiparametric spectral modeling \\ of the \\ {\it Drosophila} connectome}
\title{
\bf Semiparametric spectral modeling \\ of the {\it Drosophila} connectome
}
%\end{center}
  \medskip
\maketitle
} \fi

%\newpage

\bigskip
\begin{abstract}
\noindent 
%We continue our quest to develop principled methodologies for connectome inference in general, and for addressing the cortical column conjecture in particular
% $\dots$ but we don't say so, explicitly.
% Rather:
%We continue our quest to develop principled methodologies for network inference 
%by demonstrating the potential of interplay between graph embedding inference
%and Knowledge Base interaction
%to incorporate structure ontologies and vertex and edge attributes into the discovery of structure in (partially, errorfully) observed graphs.
We present semiparametric spectral modeling of the complete larval {\it Drosophila} mushroom body connectome.
Motivated by a thorough exploratory data analysis of the network via 
  Gaussian mixture modeling (GMM) in the adjacency spectral embedding (ASE) representation space,
  we introduce the {\it latent structure model} (LSM) for network modeling and inference.
LSM is a generalization of the stochastic block model (SBM)
  and a special case of the random dot product graph (RDPG) latent position model,
  and is amenable to semiparametric GMM in the ASE representation space.
The resulting {\it connectome code}
  % derived via semiparametric Gaussian mixture modeling composed with adjacency spectral embedding
  derived via semiparametric GMM composed with ASE
  captures latent connectome structure
  and elucidates biologically relevant neuronal properties.
\end{abstract}

\vfill

%\thispagestyle{empty}
%\newpage
%\clearpage
%\setcounter{page}{1}

%\newpage
\spacingset{1.45} % DON'T change the spacing!

\section{Introduction: Brains \& connectomes}

The term ``connectome'' was coined by \cite{3230/THESES} and \cite{10.1371/journal.pcbi.0010042}, and has come to mean any ``brain graph'';
``connectomics'' means the study of such graphs;
and ``statistical connectomics'' means the statistical analysis of such graphs.

\begin{comment}
The text in {\it italics} is from
\url{https://en.wikipedia.org/wiki/Connectome}
August 10, 2016.

{\it
A connectome is a comprehensive map of neural connections in the brain, and may be thought of as its ``wiring diagram''.
More broadly, a connectome would include the mapping of all neural connections within an organism's nervous system.
}

{\it
The production and study of connectomes, known as connectomics, may range in scale
  from a detailed map of the full set of neurons and synapses within part or all of the nervous system of an organism
  to a macro scale description of the functional and structural connectivity between all cortical areas and subcortical structures. 
The term ``connectome'' is used primarily in scientific efforts to capture, map, and understand the organization of neural interactions within the brain.

Research has successfully constructed the full connectome of one animal: the roundworm C.\ elegans (White et al., 1986,[2] Varshney et al., 2011[3]). 
Partial connectomes of a mouse retina[4] and mouse primary visual cortex[5] have also been successfully constructed. 
Bock et al.'s complete 12 TB data set is publicly available at Open Connectome Project.
}

{\it
The ultimate goal of connectomics is to map the human brain. 
This effort is pursued by the Human Connectome Project, 
 sponsored by the National Institutes of Health, 
 whose focus is to build a network map of the human brain in healthy, living adults.
}
\end{comment}

The Human Connectome Project 
 (\url{http://www.humanconnectomeproject.org})
``aims to provide an unparalleled compilation of neural data, an interface to graphically navigate this data and the opportunity to achieve never before realized conclusions about the living human brain.''
\cite{Sporns:2012:DHC:2381020} provides a recent survey of the quest to discover the human connectome.
%\url{https://mteocolphi.files.wordpress.com/2012/08/2013_sporns-connectome.pdf}
\cite{WashU2016} presents the newest results in the long history of efforts to update the Brodmann region atlas \citep{Brodmann100}.
% https://steemit.com/news/@nkdk/new-brain-map-doubles-number-of-known-regions
% http://www.smithsonianmag.com/smart-news/new-brain-map-doubles-known-number-regions-180959869/?no-ist

There are various connectomes available, 
including at the macro-scale
connectomes constructed from structural, functional, and diffusion MRI data.
%both structural and functional;
For instance, the Open Connectome Project
 (\url{http://www.openconnectomeproject.org})      % another repo = \url{http://www.humanconnectomeproject.org} ... we will eventually eat them.
 makes available connectomes collected via       % DTI, DSI, fMRI, structural MRI
  structural magnetic resonance imaging (MRI),
  functional MRI (fMRI),
  diffusion tensor imaging (DTI), and
  diffusion spectrum imaging (DSI).
  % Diffusion-weighted magnetic resonance imaging (DWI or DW-MRI) is an imaging method that ... 
  % A special kind of DWI, diffusion tensor imaging (DTI), has been used ... 
  % A variant of diffusion weighted imaging, diffusion spectrum imaging (DSI),  ...
These macro-scale connectomes are used, for example, to investigate connectivity between (sub)cortical regions.

In addition to MRI modalities,
there are behavioral connectomes (via optogenetics),
activity-based connectomes (via calcium imaging), etc.
For example,
  \cite{Activity1} and \cite{Activity2} consider activity-based connectomes, and 
  \cite{Vogelstein:2014hn} investigates a behavioral connectome obtained via optogenetic neuron manipulation.

At the neurons-as-vertices and synapses-as-edges scale,
partial connectomes of various organisms ({\it C.\ elegans}, {\it Drosophila}, zebrafish, mouse, etc.)\ are also available.
   The full connectome of the roundworm {\it C.\ elegans}
   has been made available at this micro-scale via electron microscopy
   \citep{1986RSPTB.314....1W,10.1371/journal.pcbi.1001066}.
   % (White et al., 1986,[2] Varshney et al., 2011[3]). 
This data continues to be widely studied.
Notably, there are {\it two} connectomes available for {\it C.\ elegans} --
one based on chemical synapses and one based on electrical synapses;
\cite{Chen:2016cb} presents a joint graph inference case study
of the {\it C.\ elegans} chemical and electrical connectomes.

%structural MRI -- vertices = voxels, Brodmann regions, etc ; edges = ???
%functional MRI -- vertices = Brodmann regions, etc ; edges = correlation???
%diffusion MRI (is this different?) -- vertices = voxels, Brodmann regions, etc ; edges = white matter tracks???
%activity (calcium imaging)
%behavioral (optogenetics)
%other?????
  % some based on electron microscopy
  % some based on lesion studies
  % some based on light microscopy and tracer injections.
  % some based on literature crawl.

 % Kathi: aot "This is an axon-synapse-dendrite connectome",
 % we have axo-axonic, axo-dendritic, dendro-axonic, dendro-dendritic.
 % to wit:
 % MBIN-to-KC: axo-axonic (besides the two MBINs in the calyx: axo-dendritic)
 % KC-to-MBIN: axo-axonic (besides the two MBINs in the calyx: dendro-axonic)
 % KC-to-MBON: axo-dendritic (besides the two MBONs in the calyx: dendro-dendritic)
 % KC-to-KC: axo-axonic and dendro-dendritic
 % MBIN-to-MBON: axo-dendritic
 % MBON-to-MBIN: all exist but mostly axo-dendritic and axo-axonic
 % MBON-to-MBON: all exist but mostly axo-dendritic and axo-axonic

 % {\bf connectome coding}:\\
A holy grail of connectomics is the ``connectome code'' --
  a generative model characterizing important aspects of the connectome\footnote{
Neural coding characterizes the relationship between the ongoing external environment (stimuli or behaviors) and neural activity.  
By way of analogy, connectome coding characterizes the relationship between past experience (including genetics) and neural connectivity.  
Specifically, which properties of connectomes are preserved across individuals of the same species, and which vary as a function of life history? 
Similarly, which connectome codes are preserved across species, and which are adapted to species' specific evolutionary niches?}.
This paper reports the results of a ``structure discovery'' analysis
 % {\bf structure discovery}:\\
 % f  or Jacob re ``Neuropedia Knowledge Base'':\\
 %     we won't use this terminology, but clearly the NKB was essential to our structure discovery.
  of an important first-of-its-kind
  complete
  neurons-as-vertices and synapses-as-edges electron microscopy 
  %structural circuit 
  connectome.
% of a higher-order learning center involved in memory formation or storage.
The paper is organized as follows.
Section 2 provides a brief description of our data,
  the larval {\it Drosophila} mushroom body connectome
  described in detail in \cite{EichlerSubmitted}.
Section 3 presents a thorough spectral clustering investigation of this connectome,
  demonstrating conclusively that there is one major aspect of the connectome that is insufficiently captured by this approach.
Section 4 introduces and develops the principled semiparametric spectral modeling methodology
  that we use to generate a much more satisfying connectome code for the larval {\it Drosophila} mushroom body.
%Section 5 provides ample discussion, and Section 6 presents our grandiose conclusion.
Section 5 provides both neuroscientific and methodological discussion, and Section 6 presents our conclusion.
Algorithmic details are provided in an appendix. %,
%and
%data \& code for all our analyses are available at
%% $<$\url{http://www.cis.jhu.edu/~parky/MBstr.html}$>$.
%%\\
%% %$<$\url{http://www.cis.jhu.edu/~parky/MBstructure.html}$>$.
% $<$\url{https://github.com/youngser/mbstructure}$>$.

%\newpage

\section{The larval {\it Drosophila} mushroom body connectome}

HHMI Janelia 
recently reconstructed the complete wiring diagram of the higher order parallel fiber system for associative learning in the larval {\it Drosophila} brain, 
the mushroom body (MB).
Memories are thought to be stored as functional and structural changes in connections between neurons, 
 but the complete circuit architecture of a higher-order learning center involved in memory formation or storage
 has not been known in any organism $\dots$ until now.
%\footnote{
%From {\it C.\ elegans},
%through this larval {\it Drosophila} mushroom body connectome,
%to IARPA MICrONS $\dots$ and beyond!}.
This data set provides a real and important example for initial investigation into
  synapse-level structural connectome modeling.
% structural connectome modeling at the neurons-as-vertices level.

%ontologically adaptive graph embedding,
%and
%demonstrates the potential of interplay between graph embedding inference
%and Knowledge Base interaction
%to incorporate structure ontologies 
%such as candidate motifs
%and vertex and edge attributes into the discovery of structure in observed graphs.

%The Janelia larval {\it Drosophila} mushroom body connectome
The MB connectome
was obtained via serial section transmission electron microscopy of an entire larval {\it Drosophila} nervous system
  %cite (Ohyama et al.\ 2015, Schneider-Mizell et al.\ 2016). 
  \citep{ohyama2015multilevel,schneider2016quantitative}.
This connectome contains the entirety of MB intrinsic neurons called Kenyon cells and all of their pre- and post-synaptic partners
  %cite (Eichler et al.\ 2016).
  \citep{EichlerSubmitted}.
% re "This connectome contains the entirety of MB intrinsic neurons called Kenyon cells and all of their pre- and post-synaptic partners"
% cep:
% what does the word "intrinsic" mean in "MB intrinsic neurons called Kenyon cells"?
%
% Kathi:
% It means cells that are exclusively in this neuropile and have no branches going out of this area. 
% For example MBINs and MBONs are called extrinsic for the MB, because they have branches also in other parts of the brain.
%
% cep:
% so these KCs have no other connections, except to these KCs and these PNs and these MBINs and these MBONs?
% whereas the MBINs eg have connections other than those to these neurons being studied here?

\begin{figure}[H]
\centering
\includegraphics[width=1.10\textwidth]{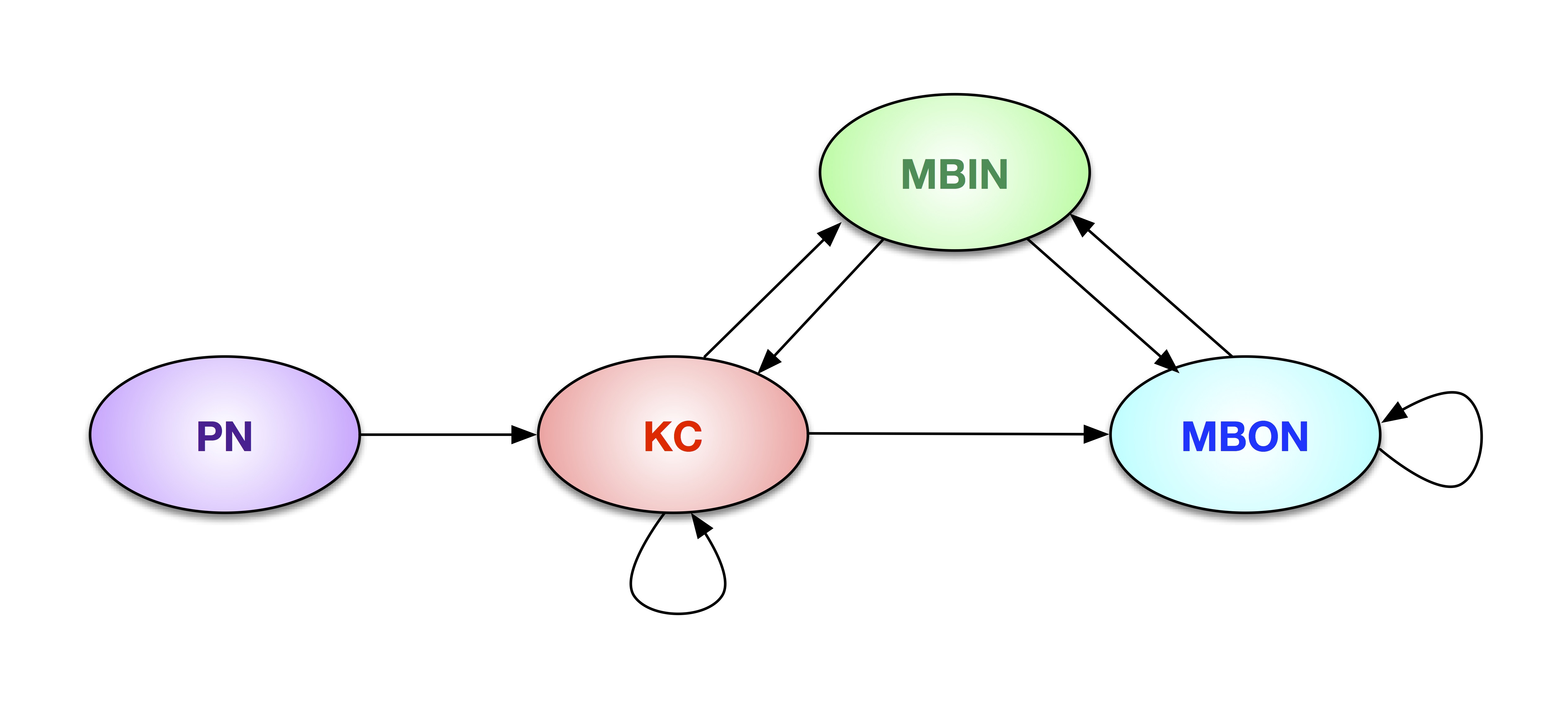}
\caption{\label{fig:Fig1} Illustration of the larval {\it Drosophila} mushroom body connectome as a directed graph on four neuron types.}
\end{figure}

\begin{figure}[H]
\centering
\includegraphics[width=1.00\textwidth]{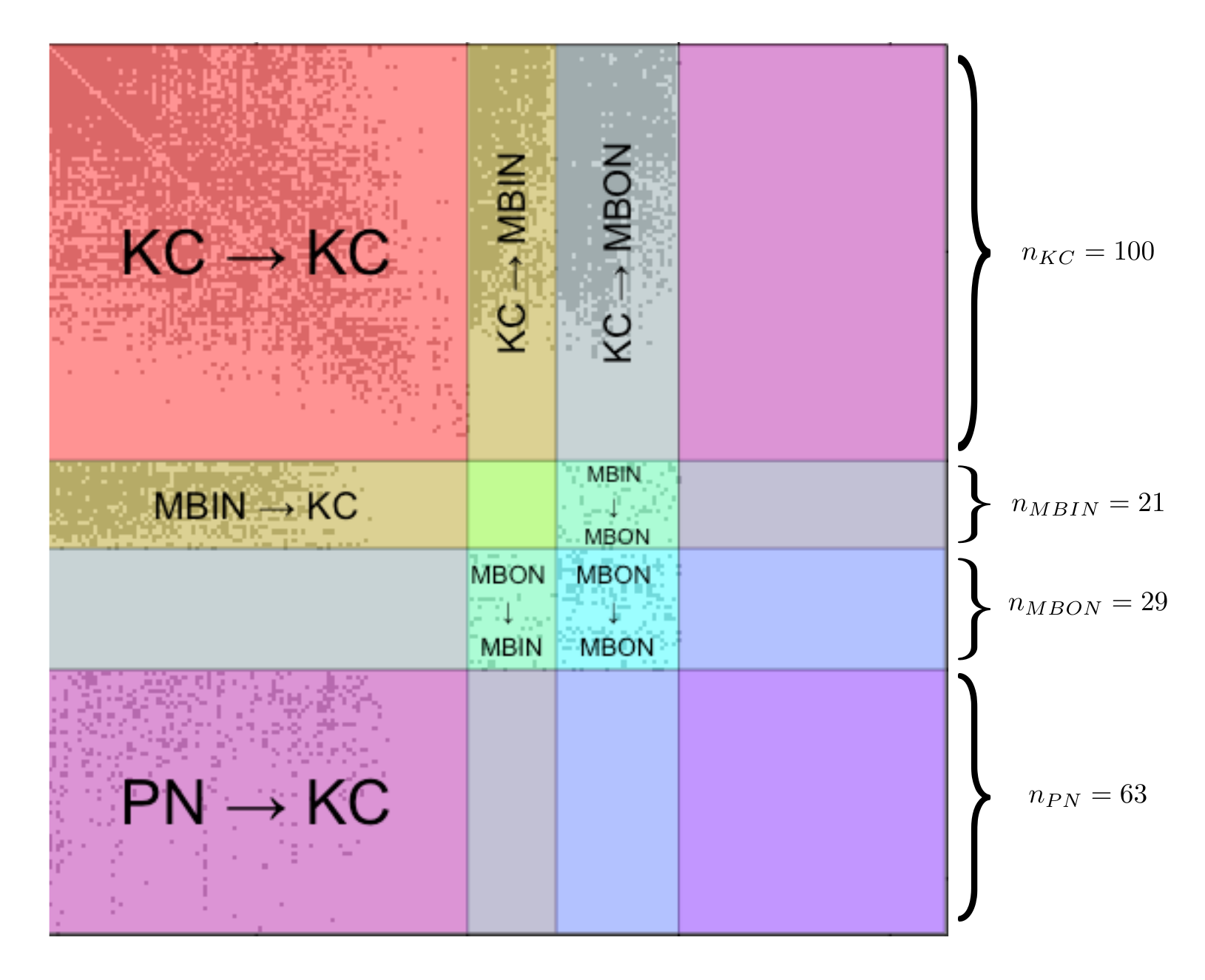}
\caption{\label{fig:Fig2} Observed data for the MB connectome as a directed adjacency matrix on four neuron types with 213 vertices 
($n_{KC} = 100$,
$n_{MBIN} = 21$,
$n_{MBON} = 29$, and
$n_{PN} = 63$)
and 7536 directed edges.
%(This data matrix is available at
% %$<$\url{http://www.cis.jhu.edu/~parky/MBstructure.html}$>$.)
% $<$\url{https://github.com/youngser/mbstructure}$>$.)
}
\end{figure}

We consider the right hemisphere MB.
 % cep:
   % what do we think about studying just one hemisphere MB in isolation?
 % Kathi: 
   % I think for this statistics approach it should be fine studying one hemisphere MB in isolation.  The reason to reconstruct both was to prove if the connections from PNs to KCs is random/ not the same on both sides or if there is previously unknown structure. 
 % cep:
   % that's fine ... but perhaps you can add some neuroscientific heft to the argument that studying just one hemisphere MB in isolation is a fine thing to do for my statistical connectomics story?
 % Kathi:
   % Well I would say exactly because the wiring is different in the two hemispheres it would be good to study the hemispheres separately. Maybe it would actually be interesting to see if the analysis of the left and right hemisphere separately shows similar results. Thus we could show that even though the wiring is random and different between the two sides (at least PN to KC, other connections are consistent between hemispheres) the overall connectomics is the same. Would that make sense?
The connectome consists of four distinct types of neurons
 -- Kenyon Cells (KC), Input Neurons (MBIN), Output Neurons (MBON), Projection Neurons (PN) --
with directed connectivity illustrated in Figure \ref{fig:Fig1}.
There are $n=213$ neurons\footnote{
There are 13 isolates, all are KC; removing these isolates makes the (directed) graph one (weakly, but not strongly) connected component with 213 vertices and 7536 directed edges.},
with
$n_{KC} = 100$,
$n_{MBIN} = 21$,
$n_{MBON} = 29$, and
$n_{PN} = 63$.
Figure \ref{fig:Fig2} displays the observed MB connectome as an adjacency matrix.
Note that, in accordance with Figure \ref{fig:Fig1},
Figure \ref{fig:Fig2} shows data (edges) in only eight of the 16 blocks.
cir

%\newpage

\section{Spectral clustering}

Due to its undeniable four-neuron-type connectivity structure,
we might think of the MB connectome,
 to first order,
 as an observation from a $(K=4)$-block directed stochastic block model (SBM) on $n$ vertices.
%{\bf citep{SBM,dirSBM}}.
(The SBM was introduced in \cite{holland1983stochastic};
 the directed version in \cite{WangWong}.)
This model is parameterized by ({\it i}) a block membership probability vector $\rho = [\rho_1,\cdots,\rho_K]$
  such that $\rho_k \ge 0$ for all $k$ and $\sum_k \rho_k = 1$
and ({\it ii}) a $K \times K$ block connectivity probability matrix $B$ with entries $B_{k_1,k_2} \in [0,1]$
governing the probability of directed edges from vertices in block $k_1$ to vertices in block $k_2$.
%See Figures \ref{fig:CircuitDiagramComplete} and \ref{fig:CircuitAdjMatComplete} for our illustrations of this model and our data,
%with blocks identified as 1 = Kenyon Cells (KC), 2 = Input Neurons (MBIN), 3 = Output Neurons (MBON), 4 = Projection Neurons (PN).
For this model of the MB connectome we have
\[B = \left[ \begin{array}{cccc}
B_{11} & B_{12} & B_{13} & 0 \\
B_{21} & 0      & B_{23} & 0 \\
0      & B_{32} & B_{33} & 0 \\
B_{41} & 0 & 0 & 0 \end{array} \right]\] 
where the 0 in the $B_{31}$ entry, for example,
indicates that there are no directed connections from any MBON neuron to any KC neuron (as seen in Figures \ref{fig:Fig1} and \ref{fig:Fig2}).

Theory and methods suggest Gaussian mixture modeling
(see, for example, \cite{mclust2012})
composed with adjacency spectral embedding
(see, for example, \cite{STFP2012}),
denoted $\mclustase$, for analysis of the (directed) SBM\footnote{
$\mclustase$ for directed SBM:
the ASE CLT \citep{Athreya:2016rc}
  suggests (mutatis mutandis) that
  concatenation of the top $K$ left/right singular vectors from a directed $K$-SBM adjacency matrix
  behaves approximately as a random sample from a mixture of $K$ Gaussians in $\Re^{2K}$.
\cite{tang_priebe2016} demonstrates that the choice between adjacency spectral embedding and Laplacian spectral embedding 
  is an empirical modeling issue as neither dominates the other for subsequent inference
%  is a sticky wicket as neither dominates the other for subsequent inference
  $\dots$ and that $K$-means is inferior to GMM for spectral clustering.
}.

Adjacency spectral embedding (ASE) of a directed graph on $n$ vertices
  (e.g., the MB connectome with $n=213$ neurons, depicted as a directed adjacency matrix in Figure \ref{fig:Fig2})
  employs the singular value decomposition (SVD)
  to represent the $n \times n$ adjacency matrix via $A = USV^{\top}$
  and chooses the top $d$ singular values and their associated left- and right-singular vectors
  to embed the graph as $n$ points in $\Re^{2d}$ via the concatenation
  $$\Xhat = \left[ U_d S_d^{1/2} ~ \rvert ~ V_d S_d^{1/2} \right] \in \Re^{n \times 2d}.$$
(The scaled left-singular vectors $U_d S_d^{1/2}$ are interpreted as the ``out-vector'' representation of the digraph,
 modeling vertices' propensity to originate directed edges;
 similarly, $V_d S_d^{1/2}$ are interpreted as the ``in-vectors''.)
Gaussian mixture modeling (GMM)
  then fits a $K$-component $2d$-dimensional Gaussian mixture model
  to the points $\Xhat_1,\cdots,\Xhat_n$ given by the rows of $\Xhat$.
If the graph is an SBM (or indeed more generally) % RDPG LPM
then $\mclustase$ provides consistent subsequent inference
\citep{STFP2012,
fishkind2013consistent,
tang2012universally, 
sussman2012universally,
lyzinski13:_perfec, 
Lyzynski2015,
Athreya:2016rc,
tang_priebe2016}.
%{\bf citep{WHAT1}}.

\begin{figure}[H]
\centering
\includegraphics[width=0.95\textwidth]{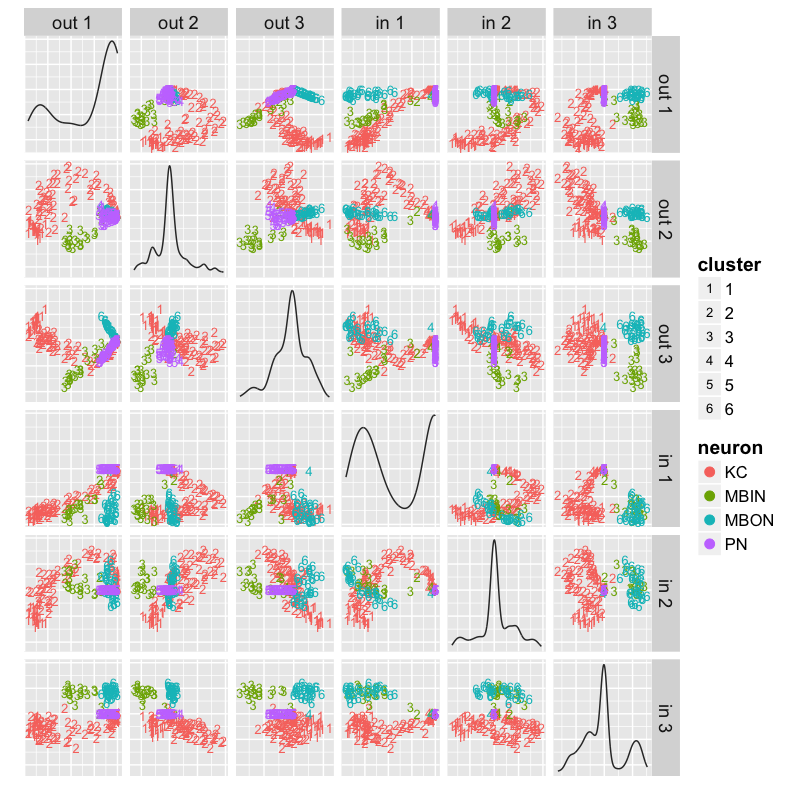}
\caption{\label{fig:MBpairs}
Pairs plot for the clustered embedding of the MB connectome 
% depicted in Figure \ref{fig:Fig2} 
into $\dhat=6$ dimensions
with $\Khat=6$ clusters.
The cluster confusion matrix with respect to true neuron types is presented in Table \ref{tab:mclust6}.
}
\end{figure}

\begin{figure}[H]
\centering
\includegraphics[width=0.95\textwidth]{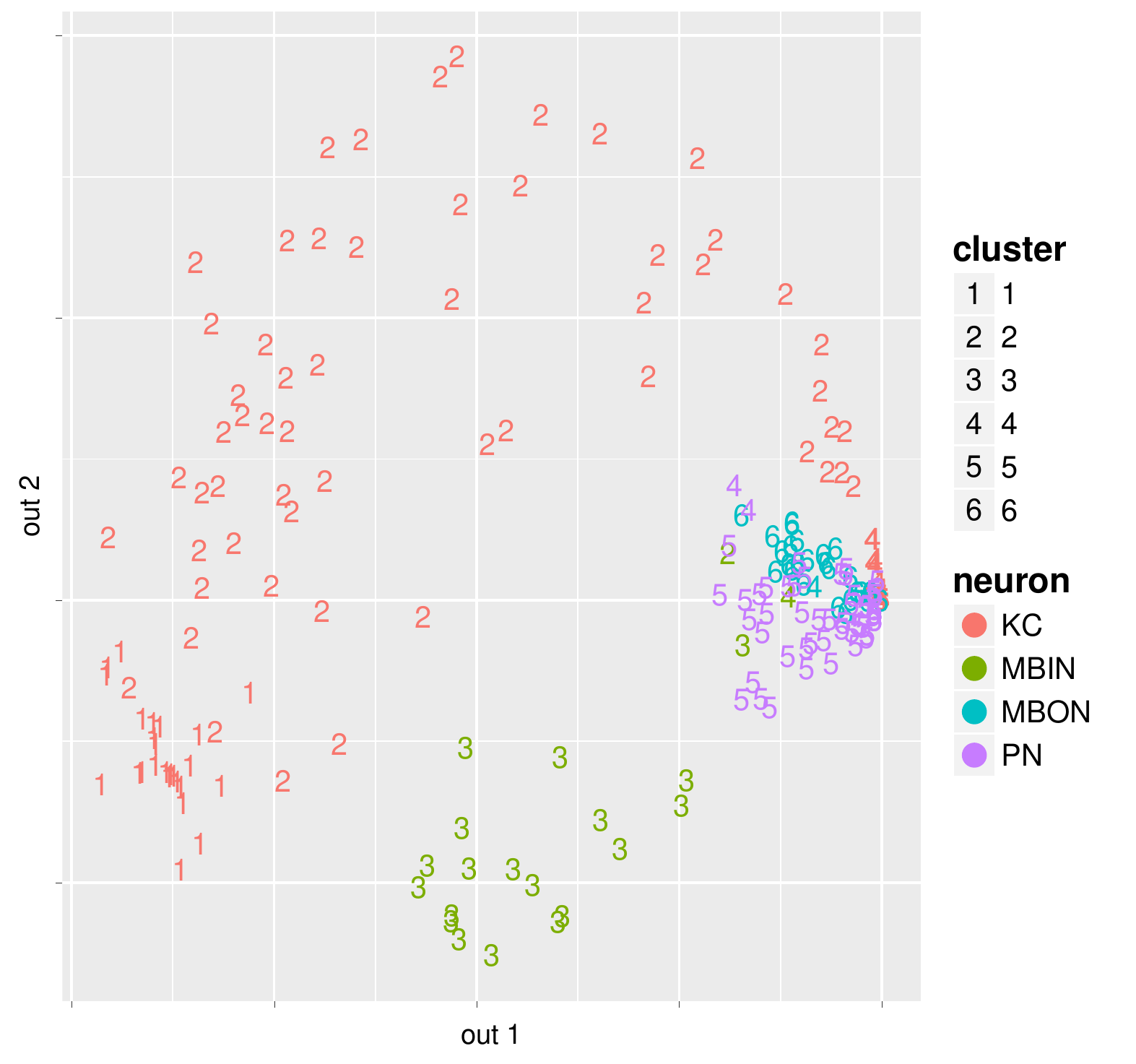}
\caption{\label{fig:MBpairs12}
Plot for the clustered embedding of the MB connectome 
% depicted in Figure \ref{fig:Fig2} 
in the out1 vs.\ out2 dimensions.
%with $\Khat=6$ clusters.
%The cluster confusion matrix with respect to true neuron types is presented in Table \ref{tab:mclust6}.
For ease of illustration,
we present embedding results in this two-dimensional subspace throughout the remainder of this manuscript.
Recall that this is a two-dimensional visualization of six-dimensional structure.
}
\end{figure}

%Adjacency spectral embedding of the digraph depicted in Figure \ref{fig:Fig2}
$\mclustase$ applied to the MB connectome
yields the clustered embedding depicted via the pairs plot presented in Figure \ref{fig:MBpairs},
with the associated cluster confusion matrix with respect to true neuron types presented in Table \ref{tab:mclust6}.
The clusters are clearly coherent with the four true neuron types.
(For ease of illustration, Figure \ref{fig:MBpairs12} presents just the out1 vs.\ out2 subspace.)

%\newpage

\newpage

There are two model selection problems inherent in spectral clustering in general,
and in obtaining our $\mclustase$ clustered embedding (Figure \ref{fig:MBpairs}) in particular:
choice of embedding dimension ($\dhat$), and choice of mixture complexity ($\Khat$).

\begin{figure}[H]
\centering
\includegraphics[width=0.400\textwidth]{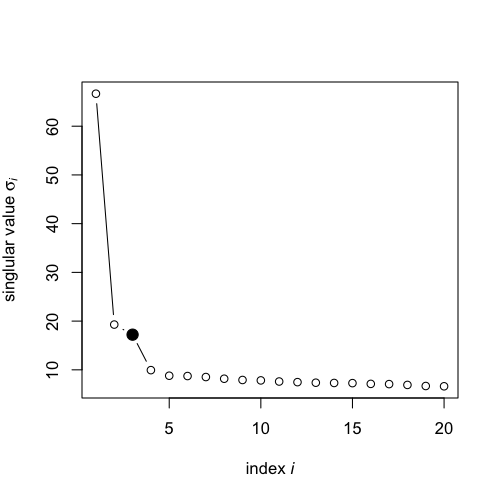}
\caption{\label{fig:MBscree}
Model Selection: embedding dimension $\dhat=6$ --
   the top 3 singular values and their associated left- and right-singular vectors --
is chosen by SVT.}
\end{figure}

% SVT
A ubiquitous and principled method for choosing the number of dimensions in eigendecompositions and SVDs
(e.g., principal components analysis, factor analysis, spectral embedding, etc.)\
  is to examine the so-called scree plot
    (the SVD scree plot for our MB connectome is presented in Figure \ref{fig:MBscree})
  and look for ``elbows'' or ``knees'' defining the cut-off between the top (signal) dimensions and the noise dimensions.
Identifying a ``best'' method is, in general, impossible, as the bias-variance tradeoff demonstrates that,
for small $n$, subsequent inference may be optimized by choosing a dimension {\it smaller than} the true signal dimension;
see Section 3 of \cite{JainDuinMao} for a clear and concise illustration of this phenomenon. %citeTrunk citeJDM.
There are a plethora of variations for automating this singular value thresholding (SVT);
Section 2.8 of \cite{Jackson} provides a comprehensive discussion in the context of principal components,
and 
\cite{chatterjee2015}
 provides a theoretically-justified (but perhaps practically suspect, for small $n$) universal SVT.
Using the profile likelihood SVT method of \cite{Zhu:2006fv} %Zhu \& Ghodsi 
yields a cut-off at three singular values, as depicted in Figure \ref{fig:MBscree}.
%(Recall that, as this is a directed graph, we have both ``in'' \& ``out'' estimated latent positions for each vertex.)
Recall that, as this is a directed graph, we have both left- \& right-singular vectors for each vertex;
thus the SVT choice of three singular values results in $\dhat=6$.

\begin{figure}[H]
\centering
\includegraphics[width=0.600\textwidth]{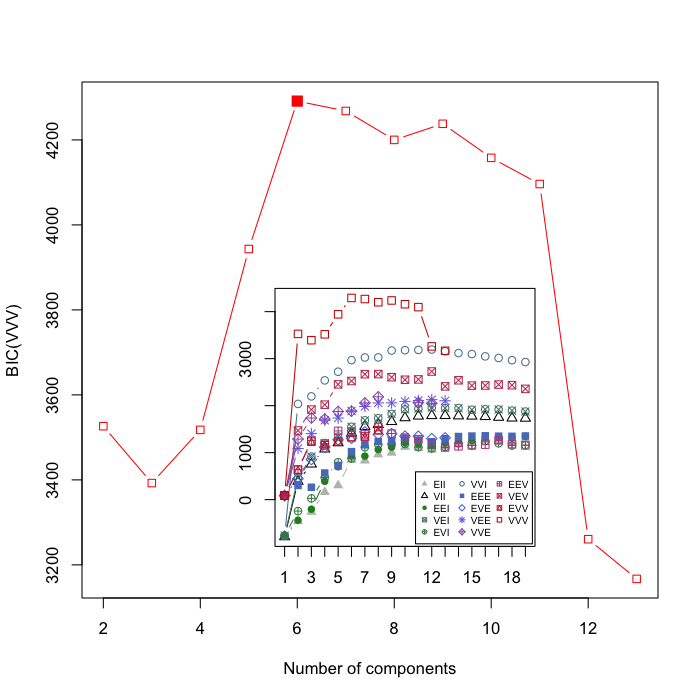}
\caption{\label{fig:MBbic}
Model Selection: mixture complexity $\Khat=6$ is chosen by BIC.
(The inset shows that the main curve -- 
BIC for dimensions 2 through 13 for MCLUST's most general covariance model, in red --
dominates all other dimensions and all other models.)
}
\end{figure}

% BIC
Similarly,
a ubiquitous and principled method for choosing the number of clusters in,
for example, Gaussian mixture models,
  is to maximize a fitness criterion penalized by model complexity.
Common approaches include
 Akaike Information Criterion (AIC) \citep{akaike1974new},
 Bayesian Information Criterion (BIC) \citep{BIC},
 Minimum Description Length (MDL) \citep{MDL},
 etc.
Again,
identifying a ``best'' method is, in general, impossible, as the bias-variance tradeoff demonstrates that,
for small $n$, inference performance may be optimized by choosing a number of clusters {\it smaller than} the true cluster complexity;
see for example \cite{Bickel} Problem 6.6.8. %e.g.\ citeBDPr688.
MCLUST's BIC \citep{mclust2012}
applied to our MB connectome embedded via ASE into $\Re^{\dhat=6}$
is maximized at six clusters,
as depicted in Figure \ref{fig:MBbic}, and hence $\Khat=6$.
(MCLUST's most general covariance model -- ``VVV'' = ellipsoidal with varying volume, shape, and orientation -- is the winner.)
%(NB: $\dhat=\Khat$ is just a coincidence.)

%\newpage

%This preliminary analysis provides the starting point for development of ontological adaptation.
%First, in keeping with the stochastic block model,
%we see from Table \ref{tab:mclust6} that this first graph embedding inference
%will report to the Knowledge Base six clusters.
The $\Khat=6$ clusters reported in Table \ref{tab:mclust6} are essentially correct, with just a few misclustered neurons
  -- e.g., cluster \#3 contains only MBIN and most of the MBIN,
           cluster \#5 contains mostly PN and most of the PN,
       and cluster \#6 contains only MBON and most of the MBON --
and, notably, KC being distributed across multiple clusters.
%The question is how will this initial structure inference be used by the Knowledge Base,
%and how will the graph inference incorporate feedback from the Knowledge Base.
%(We can easily consider the Knowledge Base as having access to edge attributes (weights, direction, etc.)
%and vertex attributes such as spatiality and even neuron type.)
%Second, of course this circuit is {\it not} a $K=4$ stochastic block model.
%How can the graph inference make use of Knowledge Base feedback concerning complex substructure, including candidate motifs?

\begin{comment}
                                cluster label
                                1  2  3  4  5  6
                          KC   25 57  0 16  2  0
                          MBIN  0  1 19  1  0  0
                          MBON  0  0  0  1  0 28
                          PN    0  0  0  2 61  0
\end{comment}

\begin{table}[H]
\centering
\begin{tabular}{lrrrrrrr}
%\begin{tabular}{lp{6em}ccccccc}
%\begin{tabular}{cccccccc}
\toprule
\addlinespace
 & & 1 & 2 & 3 & 4 & 5 & 6 \\
% \cmidrule(r){3-4}
% & $n_{\text{mc}}$ & 998 & 482 \\
 \midrule
 \addlinespace
 \text{KC}   & & {\bf 25} & {\bf 57} &  0 & {\bf 16} &  2 &  0 \\
 \text{MBIN} & &  0 &  1 & {\bf 19} &  1 &  0 &  0 \\
 \text{MBON} & &  0 &  0 &  0 &  1 &  0 & {\bf 28} \\
 \text{PN}   & &  0 &  0 &  0 &  2 & {\bf 61} &  0 \\
\end{tabular}
\caption{\label{tab:mclust6}
%Graph embedding inference $\mclustase$ for the {\it Drosophila} connectome 
$\mclustase$ for our MB connectome yields $\Khat=6$ clusters.
%This confusion matrix shows that 
The clusters are clearly coherent with but not perfectly aligned with
the four true neuron types, as presented in this confusion matrix.
%This preliminary structure estimate provides the starting point for ontological adaptation.
}
\end{table}

While BIC chooses $\Khat=6$ clusters,
 it is natural to ask whether 
 the distribution of KC across multiple clusters
 is an artifact of insufficiently parsimonious model selection.
However, choosing four or five clusters not only (substantially) decreases BIC,
%but in fact splits PN while leaving KC distributed across multiple clusters.
but in fact leaves KC distributed across multiple clusters while splitting and/or merging other neuron types.
In the direction of less parsimony,
Figure \ref{fig:MBbic} suggests that any choice from 7 to 11 clusters is competitive, in terms of BIC, with the maximizer $\Khat=6$.
Moreover, any of these choices only slightly decreases BIC,
while leaving PN, MBIN, and MBON clustered (mostly) singularly and (mostly) purely
and distributing KC across more clusters.
Tables
 \ref{tab:mclust4},
 \ref{tab:mclust5}, and
 \ref{tab:mclust7}
 show cluster confusion matrices for other choices of $K$.

\begin{table}[H]
\centering
\begin{tabular}{lrrrrr}
\toprule
\addlinespace
 & & 1 & 2 & 3 & 4 \\
 \midrule
 \addlinespace
\text{KC}    & & 26 & 	56 & 	16 & 	2 \\
\text{MBIN}  & &  0 & 	20 & 	1 & 	0 \\
\text{MBON}  & &  0 & 	28 & 	1 & 	0 \\
\text{PN}    & &  0 & 	0 & 	16 & 	47 \\
\end{tabular}
\caption{\label{tab:mclust4}
Cluster confusion matrix for $\mclustase$ with 4 clusters.
Choosing four or five clusters not only (substantially) decreases BIC (compared to $\Khat=6$),
but in fact leaves KC distributed across multiple clusters
while splitting and/or merging other neuron types.}
\end{table}

\begin{table}[H]
\centering
\begin{tabular}{lrrrrrr}
\toprule
\addlinespace
 & & 1 & 2 & 3 & 4 & 5\\
 \midrule
 \addlinespace
\text{KC}    & & 26 &	56 &	16 &	2 &	0 \\
\text{MBIN}  & & 0 &	20 &	1 &	0 &	0 \\
\text{MBON}  & & 0 &	0 &	1 &	0 &	28 \\
\text{PN}    & & 0 &	0 &	16 &	47 &	0 \\
\end{tabular}
\caption{\label{tab:mclust5}
Cluster confusion matrix for $\mclustase$ with 5 clusters.
Choosing four or five clusters not only (substantially) decreases BIC (compared to $\Khat=6$),
but in fact leaves KC distributed across multiple clusters
while splitting and/or merging other neuron types.}
\end{table}

\begin{table}[H]
\centering
\begin{tabular}{lrrrrrrrr}
\toprule
\addlinespace
 & & 1 & 2 & 3 & 4 & 5 & 6 & 7\\
 \midrule
 \addlinespace
\text{KC}    & & 25 &	42 &	15 &	0 &	16 &	2 &	0 \\
\text{MBIN}  & & 0 &	0 &	1 &	19 &	1 &	0 &	0 \\
\text{MBON}  & & 0 &	0 &	0 &	0 &	1 &	0 &	28 \\
\text{PN}    & & 0 &	0 &	0 &	0 &	2 &	61 &	0 \\
\end{tabular}
\caption{\label{tab:mclust7}
Cluster confusion matrix for $\mclustase$ with 7 clusters.
Any choice from 7 to 11 clusters only slightly decreases BIC (compared to $\Khat=6$),
while leaving PN, MBIN, and MBON clustered (mostly) singularly and (mostly) purely
and distributing KC across more clusters.}
\end{table}

%\newpage

We perform a cluster assessment to investigate the (unsupervised) selection of $\Khat=6$ via BIC and $\dhat=6$ via SVT in terms of the true neuron types.
There are numerous cluster assessment criteria available in the literature;
  we consider
  Adjusted Rand Index (ARI) \citep{hubert85},
  Normalized Mutual Information (NMI) \citep{dadiduar05},
  Variation of Information (VI) \citep{me07}, and
  Jaccard \citep{ja1912}.
(For all but VI, bigger is better; ergo, we report 1/VI for convenience.)
In Table \ref{tab:ARIetcKhat}
  we fix $\dhat=6$
  and show that BIC's $\Khat=6$ coincides with the best choice of mixture complexity.
In Table \ref{tab:ARIetcdhat}
  we find the best clustering in various embedding dimensions
  and show that SVT's $\dhat=6$ coincides with a fine choice of embedding dimension --
  choosing 4 dimensions seems approximately as good for the subsequent clustering task, while choosing 2 or 8 dimensions yields degraded performance.

\begin{table}[H]
\centering
\begin{tabular}{rrrrrrr}
\toprule
\addlinespace
  &&   ARI &	NMI &	1/VI &	Jaccard \\
 \midrule
 \addlinespace
 4 && 0.26 &  0.44 &  0.73 &  0.34 \\
 5 && 0.43 &  0.60 &  0.92 &  0.41 \\
$\Khat=6$ && {\it\bf 0.63} &	{\it\bf 0.75} &	{\it\bf 1.39} &	{\it\bf 0.57} \\
 7 && 0.55 &  0.72 &  1.15 &  0.49 \\
 8 && 0.49 &  0.68 &  0.99 &  0.43 \\
 9 && 0.49 &  0.68 &  0.96 &  0.42 \\
10 && 0.46 &  0.66 &  0.88 &  0.40 \\
11 && 0.45 &  0.65 &  0.84 &  0.39 \\
\end{tabular}
\caption{\label{tab:ARIetcKhat}
Clustering analyzed in terms of the true neuron types
(bigger is better)
shows that the (unsupervised) selection of $\Khat=6$ via BIC coincides with the objectively best clustering.
}
\end{table}

\begin{table}[H]
\centering
\begin{tabular}{rrrrrr}
\toprule
\addlinespace
          &&   ARI &	NMI &	1/VI &	Jaccard \\
 \midrule
 \addlinespace
 2        && 0.61          & 0.71          &  1.19          &  0.34 \\
 4        && 0.60          & {\it\bf 0.76} &  {\it\bf 1.43} &  0.34 \\
$\dhat=6$ && {\it\bf 0.63} & 0.75          &  1.39          & {\it\bf 0.35} \\
 8        && 0.41          & 0.67          &  0.91          &  0.30 \\
\end{tabular}
\caption{\label{tab:ARIetcdhat}
Dimension selection analyzed in terms of the true neuron types
(bigger is better)
shows that the (unsupervised) selection of $\dhat=6$ via SVT
coincides with a fine choice of embedding dimension --
choosing 4 dimensions seems approximately as good for the subsequent clustering task, 
while choosing 2 or 8 dimensions yields degraded performance.
}
\end{table}

\begin{comment}
 dim       ari       nmi       vi
   2      0.61      0.71     0.84
   4      0.60      0.76     0.70
   6      0.63      0.75     0.72
   8      0.41      0.67     1.10
\end{comment}

\begin{comment}
\begin{figure}[H]
\centering
\includegraphics[width=0.15\textwidth]{arivdhat.png}
\caption{\label{fig:arivdhat}
ARI v $\dhat$.
\\
TABLE INSTEAD!!!!!
\\
http FIGURE 20 -- we want purple curve only!
\\
Clustering analyzed in terms of the true neuron types
shows that the (unsupervised) selection of $\dhat=6$ via SVT coincides with the objectively best clustering.}
\end{figure}
\end{comment}

The conclusion of this section is that our spectral clustering of the MB connectome via $\mclustase$,
 with principled model selection for choosing embedding dimension and mixture complexity,
 yields meaningful results:
   a single Gaussian cluster for each of MBIN, MBON, and PN,
   and multiple clusters for KC.
That is, we have one substantial revision to
Figure \ref{fig:Fig1}'s illustration of the larval {\it Drosophila} mushroom body connectome as a directed graph on four neuron types:
significant substructure associated with the KC neurons.
It is a more satisfactory model of this KC substructure that we pursue in the next section.

% Our $mclust \circ ase$ model selection identifies four, five, or six clusters
%Model-based clustering 
%composed with adjacency spectral embedding ($\mclustase$) yields $\hat{K}=6$ clusters
% in the $\Re^{\hat{d}=6}$ concatenated in- and out-vector latent position estimation space
%  (see Figure \ref{fig:MBbic}.
%The associated confusion matrix is presented in Table \ref{tab:mclust6}, with Adjusted Rand Index $ARI \approx 0.63$.
%Note that in this case $\hat{K}=6$ clusters not only maximizes BIC, but also maximizes ARI.

%Interaction with neuropedia KB shows that 
%The $\hat{K}=6$ version is best (ARI):
%  three of the six clusters correspond one each to three of the neuron types (PN,MBIN,MBON),
%  and the fourth neuron type (KC) is partitioned into the other three clusters.

% Subsequent interaction with neuropedia KB then 

%\newpage 

\section{Semiparametric spectral modeling}

The spectral clustering results of the previous section --
  a single Gaussian cluster for each of MBIN, MBON, and PN,
  and from at least 3 to as many as 8 Gaussian clusters for KC --
%  and multiple clusters for KC 
% WHICH SEEM TO LIE on a ONE-D CURVE, BASED ON VIZUAL INSPECTION OF ROTATING 3D scatterplot in $\Re^6$  --
hint at the possibility of a continuous, rather than discrete, structure for the KC.

\begin{figure}[H]
\centering
\includegraphics[width=1.000\textwidth]{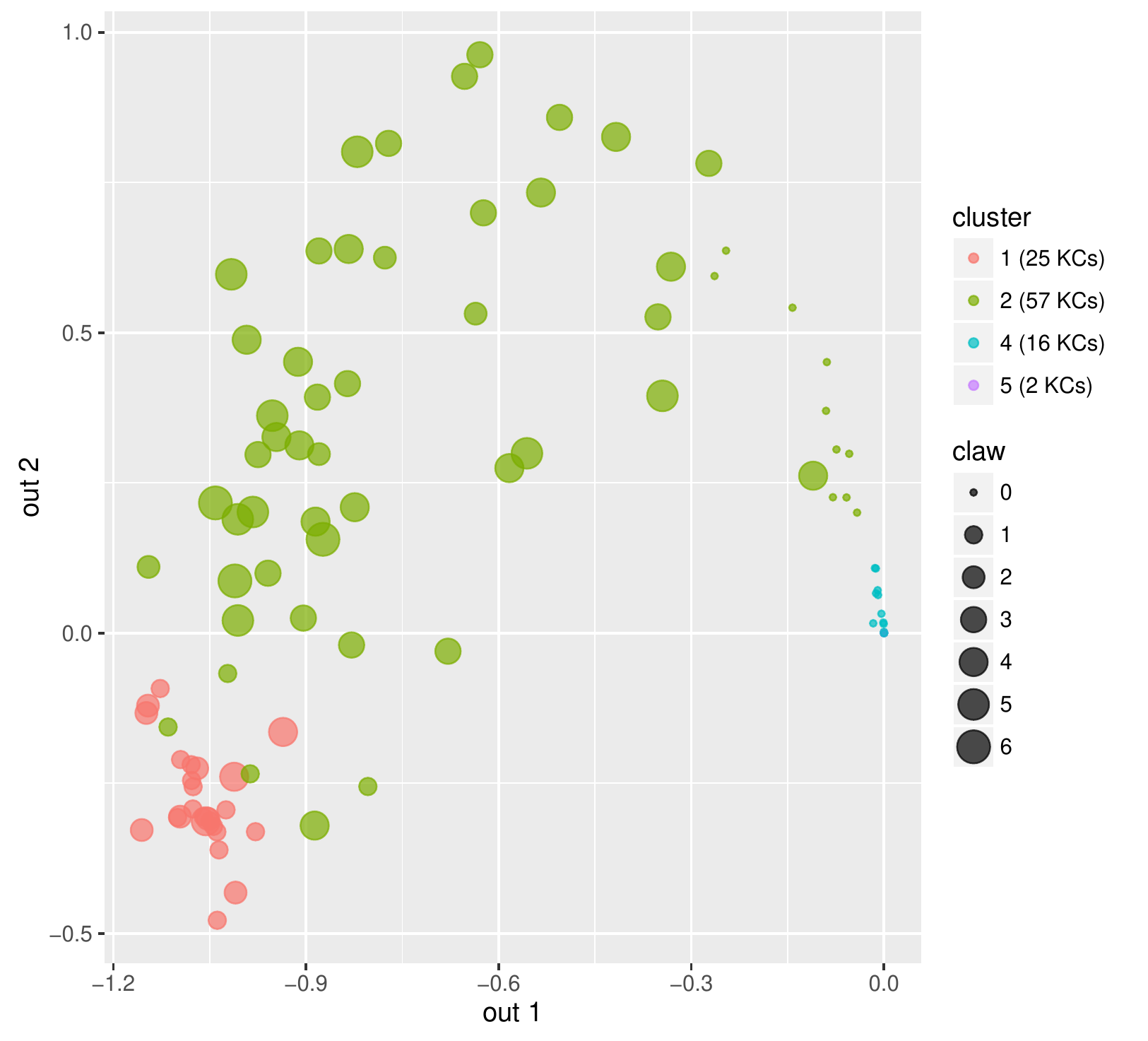}
\caption{\label{fig:MBKCageclaw}
The multiple clusters for the KC neurons are capturing neuron age.
Depicted are the first two dimensions for the KC neuron out-vectors,
  with color representing $\Khat=6$ cluster membership --
  recall from Table \ref{tab:mclust6} that the $n_{KC}=100$ KCs are distributed across multiple clusters,
  with 25 neurons in cluster \#1, 
57 in \#2,
0 in \#3,
16 in \#4,
2 in \#5, and
0 in \#6.
%\begin{table}[H]
%\centering
%\begin{tabular}{lrrrrrrr}
%\toprule
%\addlinespace
% cluster     & & 1 & 2 & 3 & 4 & 5 & 6 \\
% \midrule
% \addlinespace
% \text{KC}   & & 25 & 57 &  0 & 16 &  2 &  0 \\
%\end{tabular}
%\end{table}
The size of the dots represent the number of claws associated with the neurons.
We see from the scatter plot that the embedded KC neurons arc
  from oldest (one-claw, lower left, cluster 1, in red),
  up and younger (more claws) through cluster 2 in green, and
  back down to youngest (zero-claw, clusters 4 \& 5).
See also Table \ref{tab:MBKCageclaw}.
Recall that this is a two-dimensional visualization of six-dimensional structure.
% (This is with $\Khat=6$; perhaps one of the 7,...,11 versions is even more compelling?)
}
\end{figure}

\cite{EichlerSubmitted}
  describes so-called ``claws'' associated with each KC neuron,
  and posits that 
  KCs with only 1 claw are the oldest, followed in decreasing age by multi-claw KCs (from 2 to 6 claws), with finally the youngest KCs being those with 0 claws.
Figure \ref{fig:MBKCageclaw} and Table \ref{tab:MBKCageclaw}
use this additional neuronal information to show that the multiple clusters for the KC neurons are capturing neuron age
  -- and in a seemingly coherent geometry. % in the 6-dimensional embedding space

As the clusters for the KC neurons
  are capturing neuron age 
  -- a continuous vertex attribute --
  in a seemingly coherent geometry, % in the 6-dimensional embedding space,
%this section introduces the ``semiparametric SBM''
%this section introduces the ``latent structure model'' (LSM) generalization of the stochastic block model
%this section introduces the ``Structured SBM''
this section introduces the ``latent structure model'' (LSM) generalization of the SBM
together with the principled semiparametric spectral modeling methodology $\smclustase$ associated thereto.
Specifically,
  we fit a continuous curve to (the KC subset of) the data in latent space
  and show that traversal of this curve corresponds monotonically to neuron age.

We digress for a moment, to motivate our approach to the task at hand $\dots$
  and to develop, under the impetus of connectome modeling,
  a new direction for the theory \& methods of statistical modeling for random graphs.
  (``The wealth of your practical experience with sane and interesting problems
     will give to mathematics a new direction and a new impetus.''
     -- Leopold Kronecker to Hermann von Helmholtz (1888).)

\begin{comment}
{\it
Kathi Aug 12:
we tested the correlation between the type of KCs (single-claw, multi-claw and young) with the distance to the MB neuropile from the bundle entry point of each KC (as seen in Figure 1 i, Eichler et al.). We think that the distance to the neuropile is a measure for age, thus KCs with only one claw are the oldest (born in the embryo) followed by multi-claw KCs (2 to 6 claws) and young ones.
Also the distance to the neuropile increases with number of claws (1 to 6) and is the highest for young (0 claws) KCs. I attached a figure that shows in detail the correlation of claw number to distance to neuropile.
}
\end{comment}

\begin{table}[H]
\centering
\begin{tabular}{lrrrrrrr}
\toprule
\addlinespace
 cluster                             & & 1 & 2 & 3 & 4 & 5 & 6 \\
 \midrule
 \addlinespace
 \text{\#KCs}                           & & 25 & 57 &  0 & 16 &  2 &  0 \\
% \midrule
% \text{age: mature (1-6 claws)}     & & 25 & 47 &  --- &  0 &  0 &  --- \\
% \text{age: youngest (0 claws)}     & &  0 & 10 &  --- & 16 &  2 &  --- \\
 \midrule
 \text{claw: 1 (oldest)}            & &  15 &   4 &  --- &  0 &  0 &  --- \\
 \text{claw: 2}                     & &   7 &   4 &  --- &  0 &  0 &  --- \\
 \text{claw: 3}                     & &   0 &  15 &  --- &  0 &  0 &  --- \\
 \text{claw: 4}                     & &   3 &  13 &  --- &  0 &  0 &  --- \\
 \text{claw: 5}                     & &   0 &   8 &  --- &  0 &  0 &  --- \\
 \text{claw: 6}                     & &   0 &   3 &  --- &  0 &  0 &  --- \\
 \text{claw: 0 (youngest)}          & &   0 &  10 &  --- & 16 &  2 &  --- \\
\end{tabular}
\caption{\label{tab:MBKCageclaw}
The multiple clusters for the KC neurons are capturing neuron age via the number of claws associated with the neuron.
We see from the $\Khat=6$ clustering table, for the $n_{KC}=100$ KC neurons, that 
  cluster 1 captures predominantly older neurons,
  cluster 2 captures both old and young neurons, and
  clusters 4 \& 5 capture only the youngest neurons.
See also Figure \ref{fig:MBKCageclaw}.
% (This is with $\Khat=6$; perhaps one of the 7,...,11 versions is even more compelling?)
}
\end{table}

An alternative formulation for the directed SBM
  is as a latent position model (LPM) a la \cite{Hoff2002}.
Here one considers latent positions $X_i \sim^{iid} F$ on $\Re^d$,
  and the graph is generated from the $X_i$ via a link function or kernel
  (e.g.\ distance $d(\cdot,\cdot)$ or inner product $\langle\cdot,\cdot\rangle$).
In particular, ASE -- an SVD of the adjacency matrix --
  begs for an inner product kernel (that is, a random dot product graph, or RDPG; see e.g.\ \cite{STFP2012}).
\\

\begin{definition}[Directed Random Dot Product Graph (RDPG)]\label{def:rdpg}
Let $d_{out} = d_{in}$, and % $d=d_{out}+d_{in}$.
let $F$ be a distribution on a set $\mathcal{X} = \mathcal{Y} \times \mathcal{Z} \subset \mathbb{R}^{d_{out}} \times \mathbb{R}^{d_{in}}$
  such that $\langle y,z \rangle\in[0,1]$ for all $y\in\mathcal{Y}$ and $z\in\mathcal{Z}$.
We say that $(A,X)\sim\mathrm{RDPG}(F)$ is an
  instance of a directed random dot product graph (RDPG) if
  $X=[(Y_1,Z_1),\dotsc,(Y_n,Z_n)]^\top$ with
  $(Y_i,Z_i)\stackrel{\text{i.i.d.}}{\sim}F$, and
  $A\in\{0,1\}^{n\times n}$ is a hollow matrix satisfying
\[ \p[A|X] = \prod_{i \neq j} (Y_i^\top Z_j)^{A_{ij}}{(1-Y_i^\top Z_j)}^{1-A_{ij}}.\]
\end{definition}

The SBM as a latent position model says the latent position distribution $F$ is mixture of point masses,
  with the block membership probability vector given by the weights on these point masses
  and, for the RDPG, the block connectivity probability matrix generated by their inner products.
\\

\begin{definition}[Directed Stochastic Block Model (SBM)]
Let $d_{out} = d_{in}$, with $d=d_{out}+d_{in}$.
We say that an $n$ vertex graph $(A,X)\sim\mathrm{RDPG}(F)$
is a directed stochastic block model (SBM) with $K$ blocks if
the distribution $F$ is a mixture of $K$ point masses,
$$dF=\sum_{k=1}^K \rho_k \delta_{x_k},$$
with block membership probability vector $\vec{\rho}$ in the unit $(K-1)$-simplex % \in(0,1)^K$ satisfies $\sum_i \rho_i=1$, and
and distinct latent positions given by $x=[x_1,x_2,\ldots,x_K]^\top\in\mathbb{R}^{K\times d}$.
The first $d_{out}$ elements of each latent position $x_k$ are the out-vectors, denoted $y_k$,
and the remaining $d_{in}$ elements are the in-vectors $z_k$.
We write $G\sim SBM(n,\vec{\rho},yz^\top),$
and  we refer to $yz^\top\in\mathbb{R}^{K,K}$ as the block connectivity probability matrix for the model.
\end{definition}

The CLT for ASE of an SBM \citep{Athreya:2016rc} says that the $\Xhat_i$ 
  behave approximately as a random sample from a Gaussian mixture model (GMM)
  with the means and mixing coefficients of the mixture components providing a consistent estimate for $F$.
(An analogous CLT for Laplacian spectral embedding is available in \cite{tang_priebe2016}.)
In general, for an RDPG with any distribution $F$ on $\Re^d$ such that inner products are in $[0,1]$,
  the CLT yields approximate normality: $\Xhat_i|X_i=x_i \sim \varphi(x_i,\Sigma(x_i,F,n))$.

The above theoretical motivation is illustrated by considering
 the observed block connectivity probability matrix for our MB connectome data,
 under the assumption that it really is a 4-block directed SBM on the neuron types,
 given by

\newpage

%Bobs
%[1,] 0.36 0.45 0.49    0
%[2,] 0.38 0.00 0.12    0
%[3,] 0.00 0.09 0.21    0
%[4,] 0.08 0.00 0.00    0
\[B_{observed} \approx \left[ \begin{array}{cccc}
0.36 & 0.45 & 0.49 & 0 \\
0.38 & 0    & 0.12 & 0 \\
0    & 0.09 & 0.21 & 0 \\
0.08 & 0    & 0    & 0 \end{array} \right].\] 

\begin{figure}[H]
\centering
\includegraphics[width=1.00\textwidth]{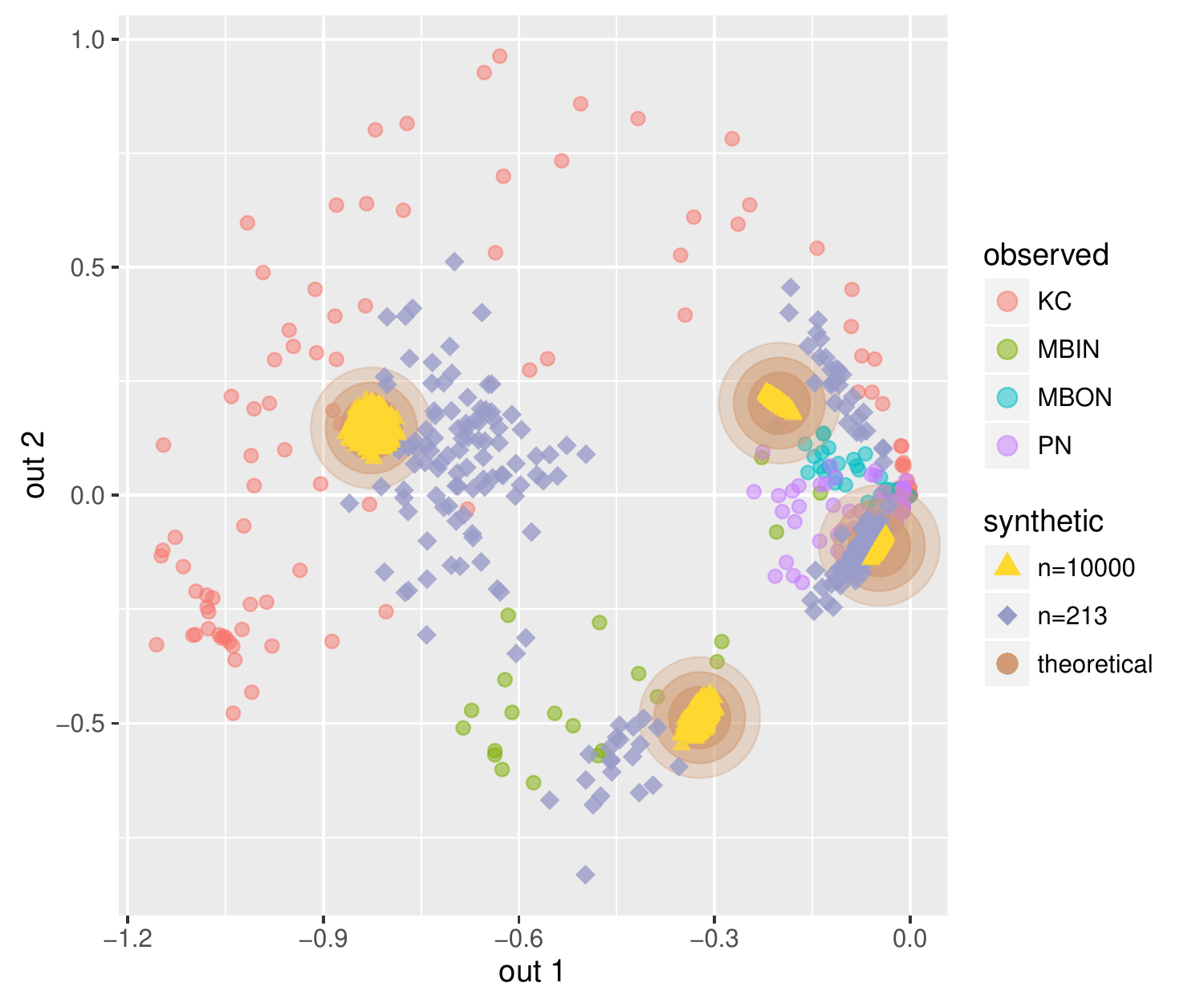}
\caption{\label{fig:MBout1out2} 
 Illustrative spectral embedding results for synthetic mushroom body directed SBM ($\synthMB$).
 %theoretical (green), large-sample (blue), small sample (red), and actual MB connectome embedding (black).
 For the observed MB connectome, the embedded MBIN, MBON, and PN are behaving as predicted,
   but clearly KC requires latent structure more complex than just $\synthMB$'s point mass.
Recall that this is a two-dimensional visualization of six-dimensional structure.}
\end{figure}

%The singular value decomposition of $B_{observed}$ 
The SVD of $B_{observed}$ provides four 4-dimensional in-vectors and four 4-dimensional out-vectors --
the true latent positions for an SBM with block connectivity probability matrix $B_{observed}$.
Together with the observed block membership probability vector
    $\rho_{observed} = [n_{KC},n_{MBIN},n_{MBON},n_{PN}]/n = [100,21,29,63]/213$,
  this yields a synthetic mushroom body directed SBM, dubbed ``$\synthMB$'',
  with latent positions $X_i \sim^{iid} F_{\synthMB}$
    where $F_{\synthMB}$ is the weighted mixture of four point masses in $\Re^8$.
These four point masses are depicted in Figure \ref{fig:MBout1out2} via brown concentric circles.
ASE provides estimated latent positions $\Xhat_i$
   for a large (10000-vertex) graph generated from $\synthMB$;
   these are depicted in gold in Figure \ref{fig:MBout1out2}
        and agree with the truth as large-sample approximation theory demands.
ASE provides estimated latent positions $\Xhat_i$
   for a small graph generated from $\synthMB$ with precisely $n=213$ vertices in precisely their observed neuron type proportions;
   these are depicted in grey diamonds in Figure \ref{fig:MBout1out2}
        and demonstrate that the large-sample approximation provides some practical guidance but has not entirely kicked in at $n=213$.
The colored circles in Figure \ref{fig:MBout1out2} depict the actual MB connectome embedding of the four neuron types;
%  one can, if so inclined, convince oneself that real embedded MBIN, MBON, and PN are behaving not-too-badly wrt $\synthMB$ $\dots$
   one can see that the real embedded MBIN, MBON, and PN are behaving approximately as expected with respect to $\synthMB$ $\dots$
   but clearly KC requires latent structure more complex than just $\synthMB$'s point mass.

All models are wrong.
We have demonstrated that the MB connectome is not a 4-block SBM,
  and this model's usefulness has been in allowing us to identify a first-order sense in which the real data deviate from the model.
The $\mclustase$ results of 
  a single cluster for each of MBIN, MBON, and PN,
  and multiple clusters indicating a geometrically coherent structure for KC
  compel us to use, as our new model, a generalization of SBM allowing KC a {\it curve}, rather than just a point, in latent space.
%We now endeavor to model the MB connectome as a 4 block $\semiparSBM$,
%   where $\semiparSBM$ denotes the ``generalized SBM'' where each ``block''
We now endeavor to model the MB connectome as a 4 component latent structure model (LSM),
   where LSM denotes the ``generalized SBM'' where each ``block''
   may be generalized from point mass latent position distribution
   to latent position distribution with support on some curve
   (with the "block" curves disconnected, as (of course) are SBM's point masses).
%So $\semiparSBM$ does have block structure $\dots$ just not quite so simple as SBM;
%and $\semiparSBM$ will exhibit clustering $\dots$ just not quite so simple as SBM.
So LSM does have structure $\dots$ just not quite so simple as SBM;
and LSM will exhibit clustering $\dots$ just not quite so simple as SBM.
Thanks to the foregoing RDPG discussion, 
  %we see that the $\semiparSBM$ is easily formulated
  we see that the LSM is easily formulated
  by considering latent position distribution $F$ 
  more general than SBM's finite mixture of point masses 
  but not arbitrary as for the fully general RDPG.
\\

\begin{definition}[Directed Latent Structure Model (LSM)] % ($\semiparSBM$)]
Let $d_{out} = d_{in}$, and 
  let $F$ be a distribution on a set $\mathcal{X} = \mathcal{Y} \times \mathcal{Z} \subset \mathbb{R}^{d_{out}} \times \mathbb{R}^{d_{in}}$
  such that $\langle y,z \rangle\in[0,1]$ for all $y\in\mathcal{Y}$ and $z\in\mathcal{Z}$.
We say that an $n$ vertex graph $(A,X)\sim\mathrm{RDPG}(F)$
is a directed latent structure model (LSM) with $K$ ``structure components'' if
%is a directed structured stochastic block model ($\semiparSBM$) with $K$ ``structure components'' if
the support of distribution $F$ is a mixture of $K$ (disjoint) curves,
$$dF=\sum_{k=1}^K \rho_k dF_k(x),$$
% $$dF=\sum_{k=1}^K \rho_k(x) I\{x \in \mathcal{C}_k\},$$
with block membership probability vector $\vec{\rho}$ in the unit $(K-1)$-simplex
and $F_k$ supported on $\mC_k$ and $\mC_1,\cdots,\mC_K$ disjoint.
%with joint latent position curves $\mathcal{C}_1,\cdots,\mathcal{C}_K$
%and probability functions $\rho_k$ such that
%  $\int_{\mathcal{X}} dF(x) = \sum_k \int_{\mathcal{C}_k} \rho_k(x) dx = 1$.
%
%The first $d_{out}$ elements of each latent position $x_k$ are the out-vectors, denoted $y_k$,
%and the remaining $d_{in}$ elements are the in-vectors $z_k$.
We write $G\sim LSM(n,\vec{\rho},(F_1,\cdots,F_K))$.
%We write $G\sim \semiparSBM(n,\vec{\rho},(F_1,\cdots,F_K))$.
%and  we refer to $yz^\top\in\mathbb{R}^{K,K}$ as the block connectivity probability matrix for the model.
\end{definition}

NB: The degree-corrected SBM \citep{MR2788206} is a special case of LSM % $\semiparSBM$
where each $\mC_k$ is a {\it ray}.

NB: The ``hierarchical stochastic block model'' (HSBM), introduced and exploited in \cite{Lyzynski2015},
   is a similarly ``generalized SBM'' where each ``block''
   may be generalized from point mass latent position distribution
   to structured latent position distribution with support given by a hierarchical mixture of points.

So now we investigate our MB connectome as an LSM % a $\semiparSBM$ 
with latent positions $X_i \sim^{iid} F$
where $F$ is no longer a mixture of four point masses with one point mass per neuron type
but instead $support(F)$ is three points and a continuous curve $\CKC$.

The approximate normality provided by the CLT for ASE of an RDPG
  compels us to consider estimating $F$ via a semiparametric Gaussian mixture model for the $\Xhat_i$'s.
Let $H$ be a probability measure on a parameter space $\Theta \subset \mathbb{R}^d \times S_{d \times d}$, % \Re^{d,d}$,
  where $S_{d \times d}$ is the space of $d$-dimensional covariance matrices,
  and let $\{\varphi(\cdot;\theta) : \theta \in \Theta \}$ be a family of normal densities.
Then the function given by
$$\alpha(\cdot;H) = \int_{\Theta} \varphi(\cdot;\theta) dH(\theta)$$
is a semiparametric GMM.
$H \in \mathcal{M}$ is referred to as the mixing distribution of the mixture,
  where $\mathcal{M}$ is the class of all probability measures on $\Theta$.
If $H$ consists of a finite number of atoms,
  then $\alpha(\cdot;H)$ is a finite normal mixture model with means, variances and proportions determined by the locations and weights of the point masses.
\cite{lindsay1983} provides theory for maximum likelihood estimation (MLE) in the semiparametric GMM.

Thus
   (ignoring covariances for presentation simplicity, so that $\theta \in \Re^d$ is the component mean vector)
 we see that the ASE RDPG CLT suggests estimating the probability density function of the embedded MB connectome
 $\Xhat_1,\cdots,\Xhat_{n=213}$,
 %under the $\semiparSBM$ assumption,
 under the LSM assumption,
 as the semiparametric GMM $\alpha(\cdot;H)$
 with $\Theta=\Re^{\dhat=6}$
 and
%   $$\alpha(\cdot;H) = \int_{\Re^{\dhat=6}} \varphi(\cdot;\theta) dH(\theta)$$
 where $H=F$ 
is supported by three points and a continuous curve $\CKC$.
%is a mixture of three point masses and one continuous curve $\CKC$.
Note that in the general case, where $\Theta$ includes both means and covariance matrices, we have $H = H_{F,n}$.
However, we emphasize that it is $F$ (or $dF$) that is the ``connectome code'' -- the key to the generative model;
 the covariances (in $dH$ but not in $dF$) that we are ``ignoring for presentation simplicity''
 are in fact nuisance parameters from the perspective of the connectome code.
The ASE RDPG CLT provides a large-sample approximation for $H_{F,n}$,
and provides a mean-covariance constraint so that
  if we knew the latent position distribution $F$ we would have no extra degrees of freedom
  (though perhaps a more challenging MLE optimization problem).
As it is, we do our fitting in the general case, with simplifying constraints on the covariance structure associated with $\CKC$.

\begin{figure}[H]
\centering
\includegraphics[width=0.95\textwidth]{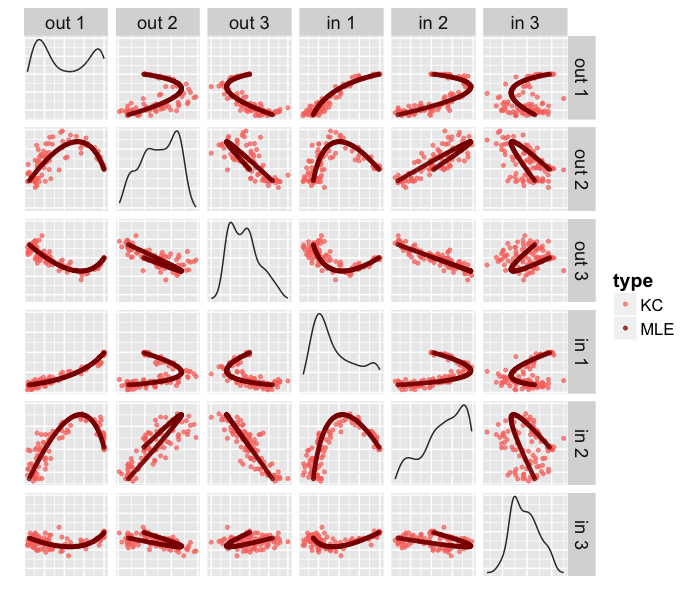}
\caption{\label{fig:YPYQ19}
Semiparametric MLE $\Chat$ for the KC latent-space curve in $\Re^6$.
%Recall that this is a two-dimensional visualization of six-dimensional structure.
}
\end{figure}

The MLE 
  (continuing to ignore covariances)
  is given by
 $$d\hat{H}(\theta) = \sum_{k=1}^3 \rhohat_k I\{\theta = \thetahat_k\} + \left(1 - \sum_{k=1}^3 \rhohat_k\right) \rhohat_{KC}(\theta) I\{\theta \in \Chat\}$$
% $$\hat{H}(\theta) = \sum_{k\in\{MBIN,MBON,PN\}}^3 \pihat_k I\{\theta = \thetahat_k\} + (1 - \sum_{k\in\{MBIN,MBON,PN\}}^3 \pihat_k) \pihat_{KC}(\theta) I\{\theta \in \Chat\}$$
 where $\thetahat_1, \thetahat_2, \thetahat_3$ are given by the means of the $\mclustase$ Gaussian mixture components for MBIN, MBON, and PN,
 and $\Chat \subset \Re^d$ is a one-dimensional curve.
%Thus, our final latent position structure discovery$^{(2)}$ consists of
%  a mixture of three discrete (Gaussian) clusters (means \& covariances)
%  together with the continuous parameterized curve (and covariance function).
Figure \ref{fig:YPYQ19} displays the MLE results from an EM optimization
for the curve $\Chat$ constrained to be quadratic, as detailed in the Appendix.
(Model testing for $\CKC$ in $\Re^6$ does yield quadratic:
testing the null hypothesis of linear against the alternative of quadratic yields clear rejection ($p < 0.001$),
while there is insufficient evidence to favor $H_A$ cubic over $H_0$ quadratic ($p \approx 0.1$).)

%\newpage

\begin{figure}[H]
\centering
\includegraphics[width=0.975\textwidth]{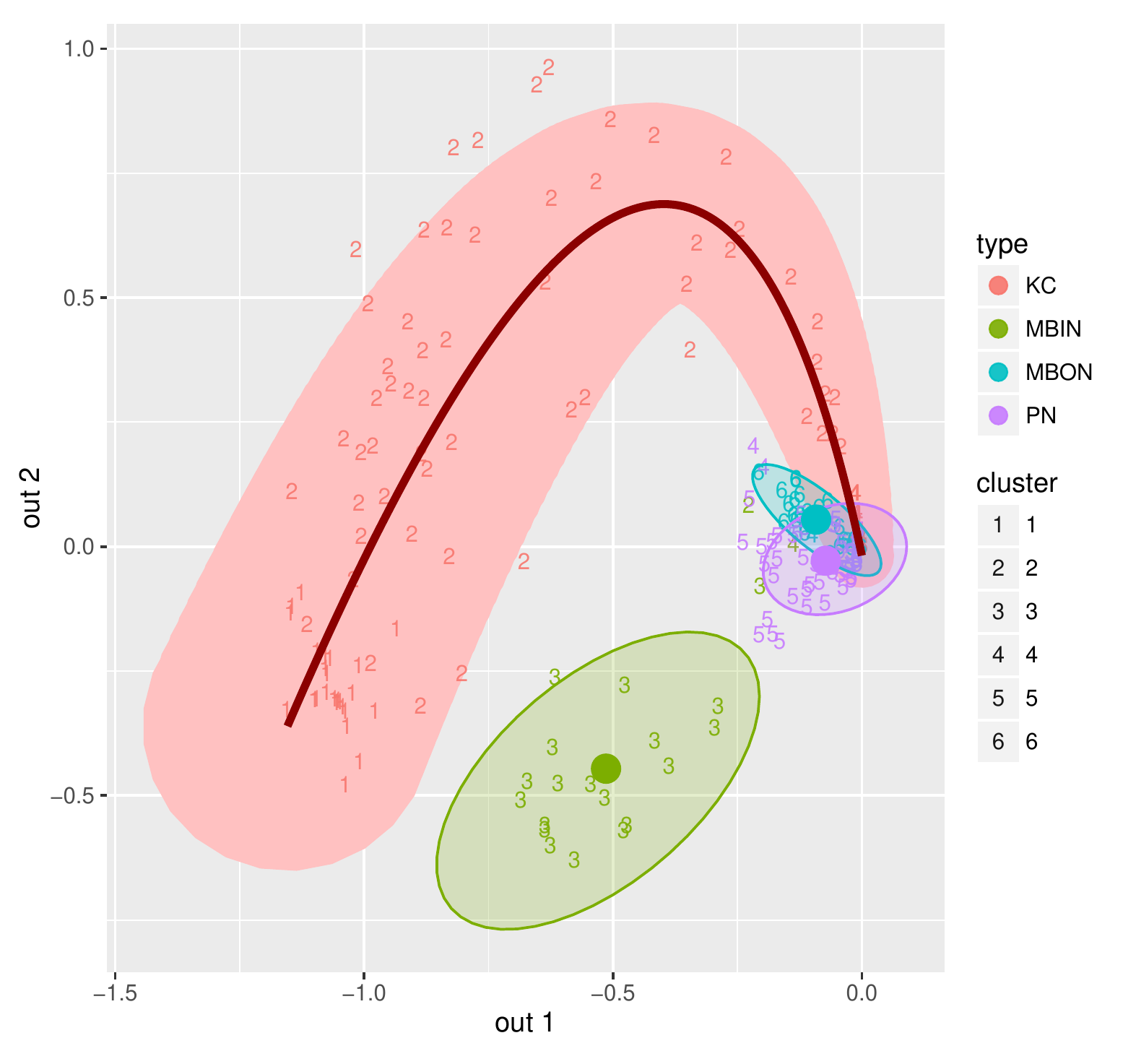}
\caption{\label{fig:MBsemiparfig}
Semiparametric spectral latent space estimate of our MB connectome as three Gaussians and a KC curve:
colors distinguish the four neuron types and
numbers distinguish the original $\Khat=6$ clusters.
%shading of the red points distinguishes KC neuron age.
Recall that this is a two-dimensional visualization of six-dimensional structure.
}
\end{figure}

That is, (continuing to ignore covariances)
  our structure discovery 
    via $\smclustase$ 
    % together with neuropedia KB interaction
  yields an $\Re^6$ latent position estimate for the MB connectome
    -- a {\it connectome code} for the larval {\it Drosophila} mushroom body --
  as a semiparametric Gaussian mixture of three point masses
and a continuous parameterized curve $\Chat$; % and its covariance function --
the three Gaussians correspond to three of the four neuron types,
and the curve corresponds to the fourth neuron type (KC) with the parameterization capturing neuron age.
See Figure \ref{fig:MBsemiparfig}.

%\newpage

\begin{figure}[H]
\centering
\includegraphics[width=0.70\textwidth]{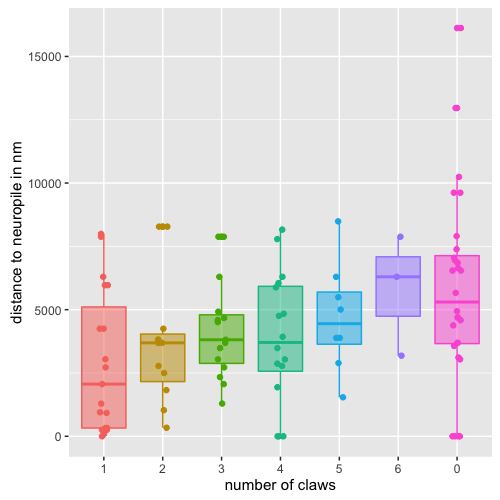}
\caption{\label{fig:KE1}
%*** Kathi's Boxplot:
Relationship between number of claws and distance $\delta_i$ (a proxy for age) for the KC neurons,
from \cite{EichlerSubmitted}.
}
\end{figure}

\begin{figure}[H]
\centering
\includegraphics[width=1.03\textwidth]{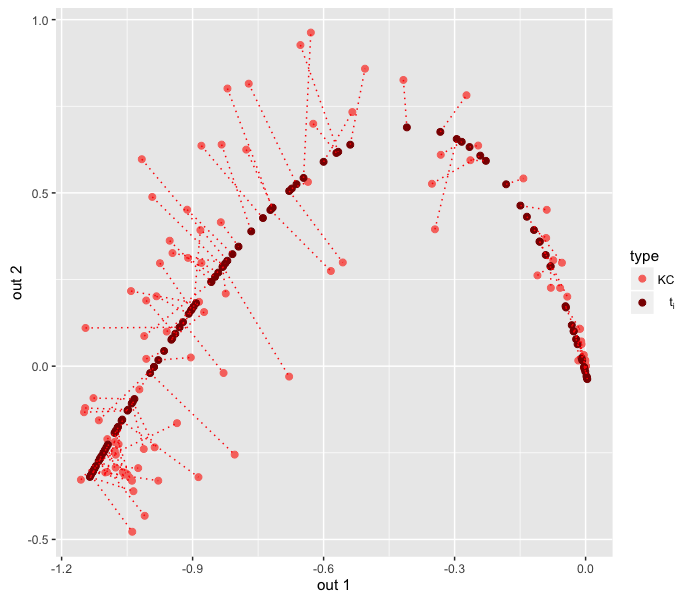}
\caption{\label{fig:YPYQ21}
Projection of KC neurons onto the quadratic curve $\Chat$, yielding projection point $t_i$ for each neuron.
Recall that this is a two-dimensional visualization of six-dimensional structure.
}
\end{figure}

\begin{figure}[H]
\centering
\includegraphics[width=1.03\textwidth]{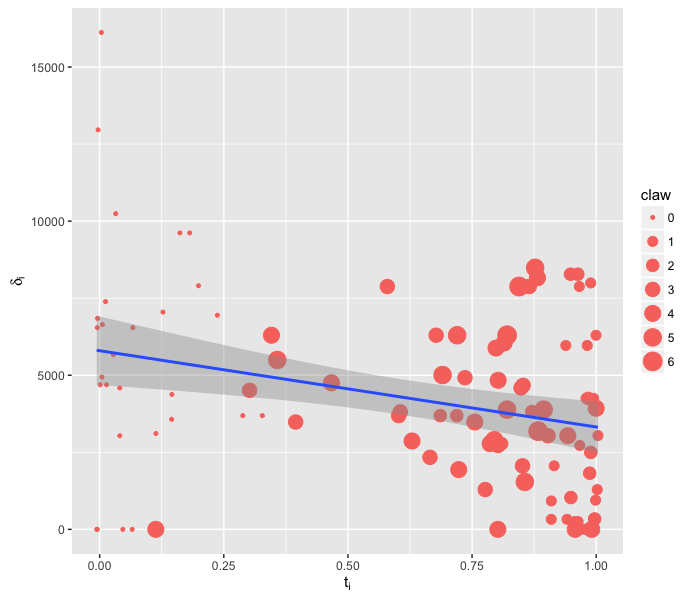}
\caption{\label{fig:YPYQ22}
The correlation between the projection points $t_i$ on the quadratic curve $\Chat$ and distance $\delta_i$ (a proxy for age) for the KC neurons
is highly significant, demonstrating that our semiparametric spectral modeling captures biologically relevant neuronal properties.
}
\end{figure}

\cite{EichlerSubmitted} suggests distance-to-neruopile $\delta_i$
 -- the distance to the MB neuropile from the bundle entry point of each KC neuron $i$ --
as a proxy for neuron age, and analyzes this distance in terms of number of claws for neuron $i$.
See Figure \ref{fig:KE1}.
%The correlation of this distance proxy for age against the projection onto the parameterized curve $\Chat$
We now demonstrate that
the correlation of this distance with the KC neurons' projection onto the parameterized curve $\Chat$
is highly significant -- this semiparametric spectral model captures neuroscientifically important structure in the connectome.
To wit,
  we project each KC neuron's embedding onto our parameterized $\Chat$
  and study the relationship between the projection's position on the curve, $t_i$, and the neuron's age through the distance proxy $\delta_i$.
\begin{comment}
       "the distance from where a KC enters the lineage bundle of KCs
       (4 per hemisphere) to the neuropile of the MB
        (which is the entry point of the last KC to enter the bundle,
       that is why some KCs have the value "0" here)."
        --- per Kathi's email on August 2nd 2016
\end{comment}
\begin{comment}
{\it
Kathi Aug 12:
we tested the correlation between the type of KCs (single-claw, multi-claw and young) with the distance to the MB neuropile from the bundle entry point of each KC (as seen in Figure 1 i, Eichler et al.). We think that the distance to the neuropile is a measure for age, thus KCs with only one claw are the oldest (born in the embryo) followed by multi-claw KCs (2 to 6 claws) and young ones.
Also the distance to the neuropile increases with number of claws (1 to 6) and is the highest for young (0 claws) KCs. I attached a figure that shows in detail the correlation of claw number to distance to neuropile.
}
\end{comment}
See Figures \ref{fig:YPYQ21} and \ref{fig:YPYQ22}.
We find significant correlation of $\delta_i$ with $t_i$ -- % is highly significant --
Spearman's $s = -0.271$, 
Kendall's $\tau = -0.205$, 
Pearson's $\rho = -0.304$,
with $p < 0.01$ in each case
-- demonstrating that our semiparametric spectral modeling captures biologically relevant neuronal properties.

%\newpage

\section{Discussion}

We briefly discuss a few of the many issues raised by this investigation:
first a few points of neuroscientific relevance,
and then some technical points regarding our methodology.

\subsection{Neuroscientific discussion points}

\subsubsection{Directed! Weighted?}
\begin{figure}[H]
\centering
\includegraphics[width=0.45\textwidth]{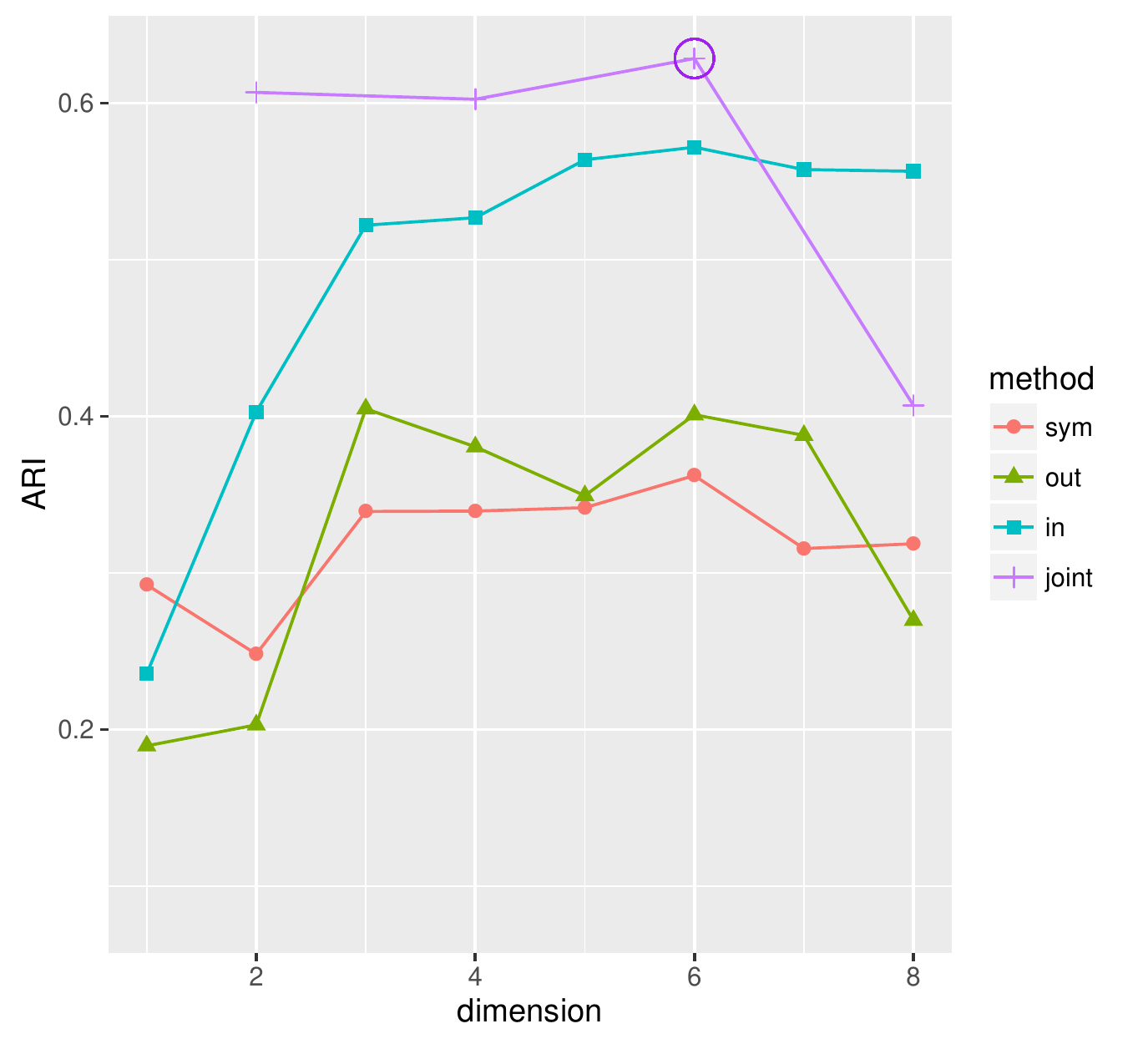}
\includegraphics[width=0.45\textwidth]{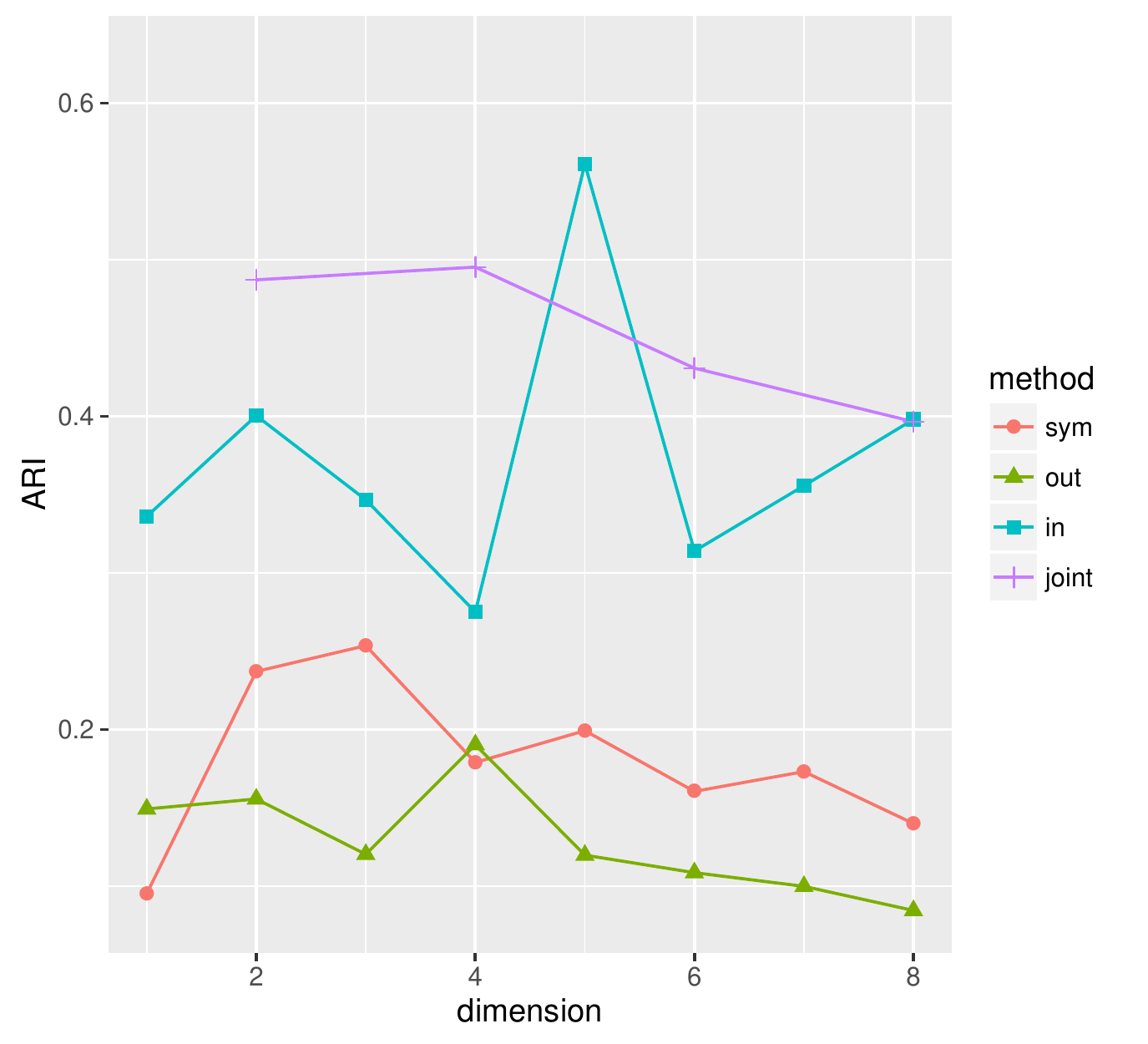}
\caption{\label{fig:arivdhatALL}
Left panel: Directed!
Plotting ARI vs.\ dimension for the $\mclustase$ analysis of the MB connectome
demonstrates that using both in- and out-vectors is superior to using in- only,
using out- only, or considering a symmetrized adjacency matrix.
That is, the directed nature of this connectome is essential to our analysis.
\\
Right panel: Weighted?
Plotting ARI vs.\ dimension for the $\mclustase$ analysis of the {\it weighted} MB connectome
demonstrates, in comparison with Figure \ref{fig:arivdhatALL}, that the best weighted version yields inferior results.
(Transformation of the weights can make the weighted results competitive.)}
\end{figure}

Inspection of 
 the six-dimensional embedding of our MB connectome 
 (Figure \ref{fig:MBpairs})
 suggests that neither the in-vectors nor the out-vectors alone suffice.
 %clustering using just the ``out'' embedding yields $ARI \approx 0.40$,
 %while using just the ``in'' embedding yields $ARI \approx 0.57$.
Figure \ref{fig:arivdhatALL} left panel demonstrates this quantitatively.

The connectome is a multi-graph -- there are multiple edges (synapses) between neurons.
We have analyzed the unweighted version.
ASE is applicable to weighted graphs, and
the analogous $\mclustase$ analysis with the weighted MB connectome yields inferior results -- see Figure \ref{fig:arivdhatALL} right panel.
 % see Figure \ref{fig:arivdhatALLw}.
NB: It does appear that one might do better using some transformation of the weights -- e.g., $w' = log(1+w)$;
this is an area of current investigation.

\begin{comment}
\subsubsection{Weighted?}
The connectome is a multi-graph -- there are multiple edges (synapses) between neurons.
We have analyzed the unweighted version.
ASE is applicable to weighted graphs, and
the analogous $\mclustase$ analysis with the weighted MB connectome yields inferior results -- see Figure \ref{fig:arivdhatALL} right panel.
 % see Figure \ref{fig:arivdhatALLw}.
NB: It does appear that one might do better using some transformation of the weights -- e.g., $w' = log(1+w)$;
this is an area of current investigation.

\begin{figure}[H]
\centering
\includegraphics[width=0.8\textwidth]{arivdhatALLw.pdf}
\caption{\label{fig:arivdhatALLw}
\end{figure}
\end{comment}

\subsubsection{Synthetic validation}

\begin{figure}[H]
\centering
\includegraphics[width=1.0\textwidth]{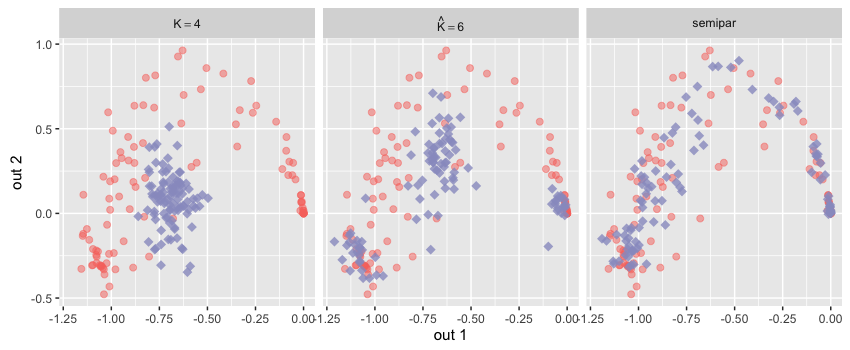}
\caption{\label{fig:synthx3}
Three synthetic KC sampling schemes, with sampled points depicted as grey diamonds and true KC embeddings depicted as red circles.
Left: the $K=4$ $\synthMB$ SBM from Section 4.
Center: the $\Khat=6$ $\mclustase$ SBM from Section 3.
Right: the $\smclustase$ estimate obtained under the 
 LSM %$\semiparSBM$ 
model from Section 4.
Recall that these are two-dimensional visualizations of six-dimensional structure.
}
\end{figure}

Figure \ref{fig:synthx3} presents synthetic KC sampling (grey diamonds) vs.\ true KC embedding (red circles).
The left panel shows the $\synthMB$ sampling from Section 4 -- just the KCs from Figure \ref{fig:MBout1out2};
sampling from the $K=4$ SBM, which models KC with a single latent space Gaussian, demonstrates the need for a more elaborate KC model.
The center panel shows the synthetic sampling using the best estimate obtained via $\mclustase$ in Section 3;
sampling from the $\Khat=6$ SBM, which models KC with three latent space Gaussians, demonstrates superiority over $\synthMB$ but still the need for a more elaborate KC model.
The right panel shows the synthetic sampling using the best estimate obtained via the $\smclustase$ methodology developed in Section 4;
here we see that sampling from the structured LSM, % $\semiparSBM$, 
which models KC with a semiparametric GMM in latent space, reproduces the KC structure amazingly well.

\subsubsection{Outlier detection and characterization}
Our semiparametric connectome code greatly facilitates the search for and characterization of outliers.
For example, there is one outlier readily apparent in 
Figure \ref{fig:YPYQ22}
 -- a KC neuron at the bottom left with small $t_i$, $\delta_i=0$, and four claws.
 % 79 KC32 #14547259   KC
(This is the larger green dot at $\approx (-0.1, 0.25)$ in Figure \ref{fig:MBKCageclaw},
  perhaps but not quantitatively an outlier without our parameterized semiparametric curve $\Chat$; 
  nothing stands out too dramatically in the four-claw boxplot in Figure \ref{fig:KE1}.)
% Kathi:
% From: "Eichler, Katharina" <eichlerk@janelia.hhmi.org>
% Subject: AW: MB structure discovery => outliers!
% Date: August 26, 2016 10:29:34 AM GMT+01:00
% To: Carey Priebe <cep@jhu.edu>, "Zlatic, Marta" <zlaticm@janelia.hhmi.org>, "Cardona, Albert" <cardonaa@janelia.hhmi.org>
Neuroscientifically, post facto investigation shows that
this neuron is clearly an outlier in the group of mature KCs,
unusual in that it has many synapses in the calyx but isn't fully grown out yet in the lobes,
explaining why this neuron might group with the 0 claw KCs having small $t_i$ in Figure \ref{fig:YPYQ22}.

This result serves as both
an example of the utility of our theory and methods for subsequent neuroscientific investigations
and
an empirical validation of our claim that
   the LSM %$\semiparSBM$ model 
and the associated $\smclustase$ estimation methodology capture biologically relevant neuronal properties in the MB connectome.

\subsubsection{Hemispheric validation: right vs.\ left}
\begin{figure}[H]
\centering
\includegraphics[width=1.0\textwidth]{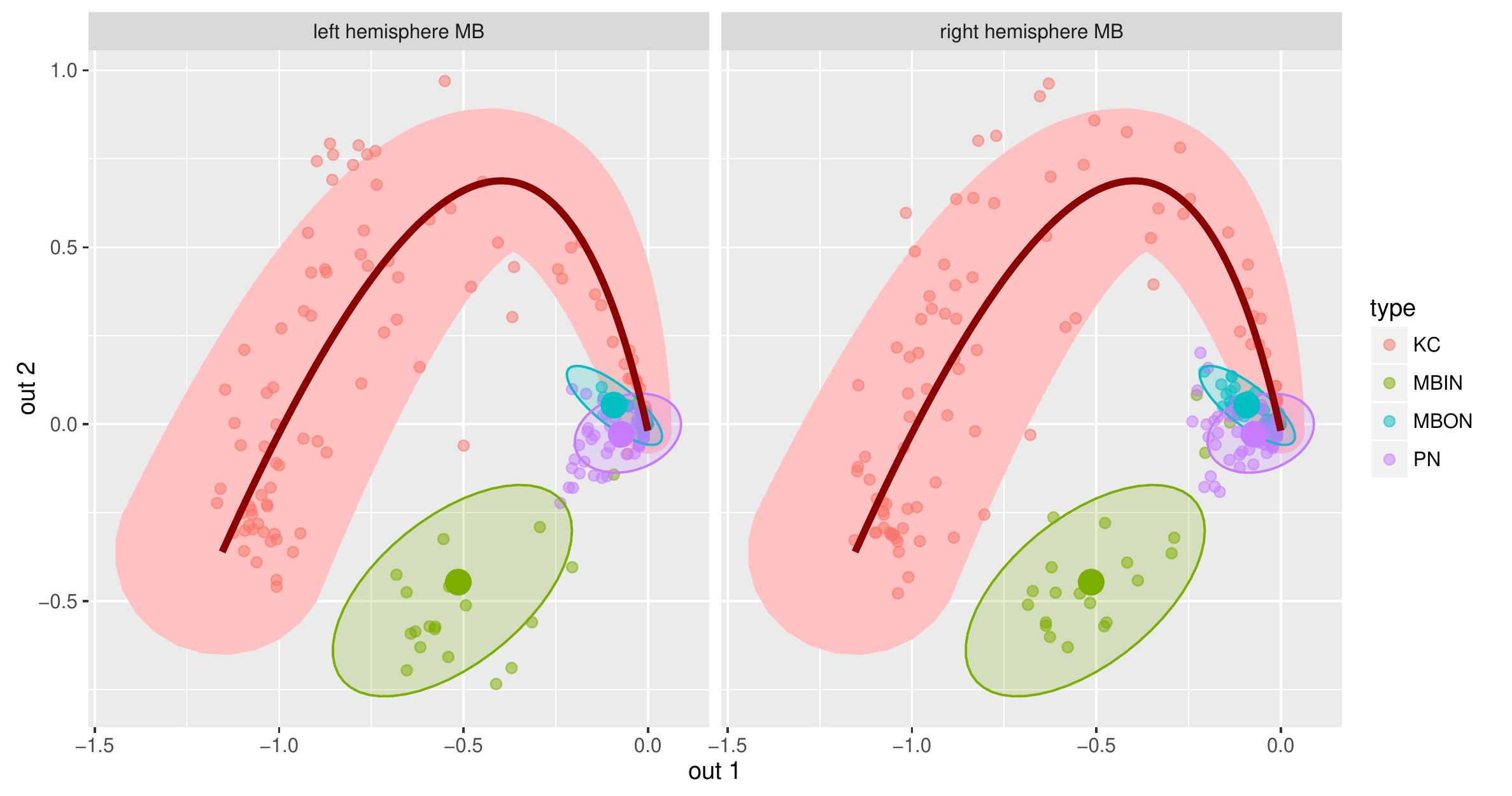}
\caption{\label{fig:leftright}
Right: the right hemisphere MB data and $\smclustase$ estimate; the structure is captured.
Left: the left hemisphere MB data superimposed on the right hemisphere MB estimate; the fit is compelling.
Recall that these are two-dimensional visualizations of six-dimensional structure.
}
\end{figure}

We have considered the right hemisphere MB. 
In the absence of $m>1$
  -- that is, MB connectomes for other larval {\it Drosophila} animals, which data are not yet available --
  we compare and contrast
  our estimate obtained on the right connectome with data from the left connectome.
(Indeed, we developed the theory \& methods and the $\smclustase$ estimate for the right hemisphere MB
without ever looking at the left hemisphere MB data $\dots$ for just this validation purpose.)
Figure \ref{fig:leftright} shows that the $\smclustase$ right hemisphere MB estimate (right panel -- a repeat of Figure \ref{fig:MBsemiparfig})
not only captures the structure in the right connectome,
but also provides compelling evidence (left panel) that the same right connectome estimate 
 captures the structure in the left connectome data as well.
(NB: The analogous result holds when we estimate using the left connectome -- the left estimate captures the right structure.)

\subsection{Methodological discussion points}

Our work is similar in spirit to the pioneering ``color circle'' perception work of \cite{ekman_54},
the ``horseshoes'' structure discovery of \cite{Horseshoes}, etc.

It is established \citep{tang_priebe2016} 
that there is no uniformly best choice between adjacency spectral embedding (ASE) and Laplacian spectral embedding (LSE) for spectral clustering.
The extension of ASE to directed graphs is straightforward (as we have seen), 
 while LSE for directed graphs remains an open area of investigation (see, e.g.,\ Section 4.3.2 in the recent survey by \cite{Malliaros201395}).
% http://www.sciencedirect.com/science/journal/03701573/533
% Section 4.3.2
% Here we will present extensions to directed networks and more precise, we will examine the generalization of the Laplacian matrix for directed graphs.

It is established \citep{tang_priebe2016} 
that $K$-means is inferior to GMM for spectral clustering.
That is, however, a limit theorem; it says little about our real MB connectome with $n=213$ vertices.
Therefore, we did consider replacing GMM with $K$-means in the Section 3 investigation, everything else remaining the same;
the results were that we did still get $\Khat=6$, but the clustering solution was evidently much less consistent with the true neuron types,
and ARI was significantly degraded (0.42 with $K$-means vs.\ 0.63 with GMM).

In practice, for small $n$, it is empirically useful to augment the diagonal of the adjacency matrix
(default: $degree(v)/(n-1)$ for undirected graphs) prior to ASE.
As a general rule, we use this augmentation.
For thoroughness, we did consider $\mclustase$ {\it without} diagonal augmentation;
the results were that we did still get $\dhat=6$, but the clustering solution was evidently much less consistent with the true neuron types,
and ARI was significantly degraded (0.36 without diagonal augmentation vs.\ 0.63 with).

\section{Conclusion}

%connectome coding:
%i very much like your overarching goal in the first paragraph,
%  "reveal the principles of neural circuits across [...] scales"
%and
%  "to reveal these principles and their mechanisms",
%and the last paragraph's
%  "to discover the principles of connectome coding".

In a recent PNAS opinion piece \citep{GGPNAS2016},
Donald \& Stuart Geman
describe
the `usual explanation' for a perceived lack of `fundamental innovation'
in areas such as brain science:
that such systems are somehow `inherently too complex,' `unsimplifiable,' `not amenable to abstraction.'
The quest for a connectome code
that reveals the principles and mechanisms of the connectivity of neural circuits
%and C5's proposed development and deployment of an enabling combined technology and theory hub,
seems to be in keeping with their position that this `usual explanation' may be shortsighted.

Motivated by the results of a spectral clustering investigation of
  the recently-reconstructed synapse-level larval {\it Drosophila} mushroom body structural connectome,
  which demonstrate conclusively that modeling the Kenyon Cells (KC) demands additional latent space structure,
  we have developed semiparametric spectral modeling.
Exploratory data analysis suggests that the MB connectome can be productively approximated by a 
 % 4 block $\semiparSBM$,
 4 component latent structure model (LSM),
  and the resulting 
MB connectome code 
derived via $\smclustase$
captures biologically relevant neuronal properties.

Of course, the true connectome code is more elaborate,
and cannot be completely encompassed by any simple latent position model --
such a model precludes the propensity for transitivity, e.g.\ --
but our semiparametric spectral modeling provides another step along the path.
In terms of a (partial) ladder of biological scales
  -- e.g., {\it C.\ elegans}, {\it Drosophila}, zebrafish, mouse, primate, and humans --
this works moves us off the first rung for analysis of a complete
neurons-as-vertices and synapses-as-edges connectome.

Next steps include extending this investigation to (currently unavailable) data for
(a) new complete synapse-level larval {\it Drosophila} MB structural connectomes from different animals,
(b) new complete supersets of the synapse-level larval {\it Drosophila} MB structural connectome,
and
(c) new complete synapse-level structural connectomes from different species, such as the adult {\it Drosophila}.
Furthermore,
mutli-modal connectome analyses, combining 
complete synapse-level structural connectomes with other modalities such as
behavioral connectomes obtained via optogenetics and activity-based connectomes obtained via calcium imaging,
promise additional advances in connectome coding.

\section*{Appendix:
Constrained Maximum Likelihood Estimation of the Semiparametric Gaussian Mixture Model for Adjacency Spectral Embedding of a Latent Structure Model}

The CLT for the ASE of an RDPG \citep{Athreya:2016rc} gives that the $\Xhat_1,\cdots,\Xhat_n$
are approximately normal about the true (unobserved) latent positions $X_1,\cdots,X_n$.
Of course, this is a large-sample (asymptotic) result;
furthermore, the theory does {\it not} provide mutual independence of all $n$ embedded points,
but rather of only a fixed finite $n'$ of the $\Xhat_1,\cdots,\Xhat_n$, as $n \to \infty$.
Nevertheless, we proceed assuming independence with what we call ``MLE in the embedding space''.

We model the MB connectome embedding
$\Xhat_1,\cdots,\Xhat_n \in \mathbb{R}^d$
as a semiparametric GMM
\begin{align*}%\label{eq:semipar_mixture}
\alpha(\cdot;H) =\int_{\Theta} \varphi(\cdot;\theta) dH(\theta)
\end{align*}
where $\{\varphi(\cdot;\theta) : \theta \in \Theta \}$ is the family of $d$-dimensional Gaussian densities,
with $\theta=(\mu,\Sigma)$, $\mu \in \mathbb{R}^d$ and $\Sigma \in S_{d \times d}$;
$H$ represents a probability measure on the parameter space $\Theta \subset \mathbb{R}^d \times S_{d \times d}$.

Our exploratory data analysis presented above suggests we consider $dH$ to be a four component mixture model,
 with the first three components being point masses
 and a final component (for the KC neurons) having continuous support on a one-dimensional curve $\mathscr{C}_{KC}$:
\begin{align*}
dH(\mu, \Sigma) = \sum_{k=1}^{3} \rho_k I\{\mu=\mu_k, \Sigma=\Sigma_k \} +
                  (1-\sum_{k=1}^{3} \rho_k ) \rho_{KC}(\mu,\Sigma) I\{ (\mu,\Sigma) \in \mathscr{C}_{KC}\}
\end{align*}
where $\int_{\mathscr{C}_{KC}} \rho_{KC}(\mu,\Sigma) = 1$ and $\mathscr{C}_{KC}$ is a curve in $\mathbb{R}^d \times S_{d \times d}$.  
%Note that we could write $\rho_{KC}(\mu,\Sigma)$ but because of mean-variance constrains, there is a one-to-one mapping between them, hence we simplify it to $\rho_{KC}(\mu)$.
(Note that the ASE CLT provides a mean-variance constraint which we do not attempt to take advantage of here.)

%\newpage

For the remainder of this appendix we focus on the final non-trivial component in $dH$,
and consider
\begin{align*}
dH = \rho_{KC}(\mu,\Sigma) I\{ (\mu,\Sigma) \in \mathscr{C}_{KC}\}
\end{align*}
and the semiparametric MLE
\begin{align*}%\label{eq:MLE}
\widehat{H}=\arg\max_{H \in \mathcal{M}} \sum_{i=1}^n \log \alpha(\Xhat_i; H)
\end{align*}
over a space of constrained probability measures $\mathcal{M}$ described below.

This optimization for $\widehat{H}$ has been widely studied.
%The properties of the above optimization is not guaranteed because of the infinite dimensionality of $H$.  
%There have been extensively research on properties of such an estimation.  
\citet{Kiefer1956} first established the consistency of the semiparametric MLE under suitable conditions. 
%However, the computation of such an MLE remained unresolved until \citet{Laird1978} and \citet{lindsay1983}.  
However, many issues remained unresolved until \citet{Laird1978} and \citet{lindsay1983}.  
The work of \citet{lindsay1983} has established that there indeed exists a maximizer $\widehat{H}$
  and provides conditions under which its uniqueness can be established;  
  the nature of the maximizer $\widehat{H}$ is further revealed to be that of a discrete distribution with finitely many mass points.  
Moreover \citet{lindsay1983} shows that the number of mass points $K$
  (which is random, depending upon the sample $\Xhat_1, ... ,\Xhat_n$)
  is bounded above by the number of distinct observations in the sample;  
  that is, $1 \leq K(\Xhat_1, ... ,\Xhat_n) \leq n$.
%The results from \citet{lindsay1983} 
These results 
  indicate that simply applying the EM algorithm will serve the purpose of estimating $H$
  because the maximizer 
  %of \eqref{eq:MLE} 
  $\widehat{H}$
  is essentially a finite GMM.
(Due to the limitations of the EM algorithm, the computational speed for such an estimation can be a challenge. 
Alternative computational approaches include
  Intra Simplex Direction Method (ISDM) by \citet{Lesperance1992},
  Generalized ISDM by \citet{Susko1998}, and
  Constrained Newton Method (CNM) by \citet{Wang2010}.
For a comprehensive review of semiparametric MLE for mixture models, see \citet{Lindsay1995}.)

Based on these previous results, 
  and after settling on a value for $K$,
  we can directly fit a $K$-component Gaussian mixture to the data.  
To facilitate such an estimation and reduce the complexity of the optimization, 
{\it and motivated by the MB connectome exploratory data analysis presented in the main body of this paper},
we further assume that $H$ is supported on
\begin{align*}
\mathscr{C}_{KC}=\{(\mu(t),\Sigma(t)):  &\mu(t) = (1-t)^2 m_1 + 2 t (1-t) m_2 + t^2 m_3, \\
&\Sigma(t) = ((1-t) \sigma^2_1 + t \sigma^2_2) I_{d \times d}, \\
&0 \leq t \leq 1\}
\end{align*}
where  $m_1, m_2, m_3 \in \mathbb{R}^d$, $\sigma^2_1, \sigma^2_2 \in \mathbb{R}^+$ and $I_{d \times d}$ is the identity matrix.  
For $0 \leq t \leq 1$,
  $\mu(t)$ represents a quadratic curve in $\mathbb{R}^d$ and
  %$\Sigma(t)$ represents a monotonically changing diagonal matrix.  
  $\Sigma(t)$ represents a linear structure for the covariance;
thus, $\mathscr{C}_{KC}$ is a curve in $\mathbb{R}^d \times S_{d \times d}$.  
This parametrization has significantly simplified the structure of $H$.
Therefore, 
after settling on $K=7$ for the KC neurons based on a BIC analysis analogous to that presented in Figure \ref{fig:MBbic}
  but accounting for our semiparametric constraints,
we fit the following $K$-component GMM to the data:
\begin{align*}
g(\cdot)=\sum_{j=1}^{K} \pi_j \varphi\big(\cdot ; \mu_j, \Sigma_j\big),
\end{align*}
where $(\mu_j, \Sigma_j)$ are equally spaced on the curve $\mathscr{C}_{KC}$. 
Since such a Gaussian mixture restricts its means and variances to satisfy the support of $H$,
  the estimation can be easily performed using the EM algorithm.

%\newpage

We propose the following EM algorithm to estimate such a mixture model.
\begin{enumerate}
\item
Given initial values: $\{ m_1, m_2, m_3 \in \mathbb{R}^d, \sigma^2_1, \sigma^2_2 \in \mathbb{R}^+, \pi \in \mathbb{R}^K \}$.

\item
E step: compute the conditional expectation of class labels
\begin{align*}
z_{ij} = \frac{ \pi_j \varphi(x_i, \mu_j ,\Sigma_j) }{ \sum_{j=1}^{K} \pi_j \varphi(x_i, \mu_j ,\Sigma_j) }
\end{align*}
where $(\mu_j, \Sigma_j) \in \mathscr{C}_{KC}$.
\item
M step: maximize the complete likelihood $\mathcal{L}^c=\sum_{i=1}^{n}\sum_{j=1}^{K} z_{ij} \log \big( \pi_j \varphi(x_i, \mu_j ,\Sigma_j)\big)$ and obtain the parameter estimates
\begin{align*}
\hat{\pi}_j = \frac{1}{n} \sum_{i=1}^{n} z_{ij} \qquad \mbox{and} \qquad \{ \hat{m}_1, \hat{m}_2, \hat{m}_3, \hat{\sigma}^2_1, \hat{\sigma}^2_2 \} = \underset{m_1, m_2, m_3, \sigma^2_1, \sigma^2_2}{\arg\max} \mathcal{L}^c
\end{align*}
such that $\mu_j=\mu\big((j-1)/(K-1)\big)$ and $\Sigma_j=\Sigma\big((j-1)/(K-1)\big)$ for $j=1,...,K$.  

\item
Use the estimates from Step 3 as the initial values, and repeat Steps 2 and 3 until convergence.

\item
Let
\begin{align*}
\widehat{\mathscr{C}}_{KC} &= \{(\mu(t),\Sigma(t)): \mu(t) = (1-t)^2 \hat{m}_1 + 2 t (1-t) \hat{m}_2 + t^2 \hat{m}_3, \\
& \qquad \qquad \Sigma(t) = ((1-t) \hat{\sigma}^2_1 + t \hat{\sigma}^2_2) I_{d \times d}, \\
& \qquad \qquad 0 \leq t \leq 1\},\\
\hat{\rho}(\mu(t),\Sigma(t))&=\sum_{j=1}^K \pi_j I \{ \mu(t)=\hat{\mu}_j,\Sigma(t)=\hat{\Sigma}_j \}.
\end{align*}
%where $\hat{\mu}_j$ are equally space in $\widehat{\mathscr{C}}_{KC}$.
\end{enumerate}

%Note that in Step 3 the optimization with respect to the $\pi_j$ and the other parameters are orthogonal, and hence can be conducted separately.

We apply the proposed EM algorithm on the adjacency spectral embeddings of $n=100$ KC neurons.  
The results are presented in Figure \ref{fig:YPYQ19} in the manuscript, where the bold curve represents $\widehat{\mathscr{C}}_{KC}$.  
% In addition, we plot our estimated $\hat{\rho}(\mu)$ in Figure \ref{fig:rho}

\citet{Kiefer1956} established the consistency of the semiparametric MLE $\widehat{H}$.
%in \eqref{eq:MLE}.  
Since our estimation is conducted in a restricted parameter space $\mathscr{C}_{KC}$, 
as long as the true parameter is inside this restricted parameter space the consistency of our estimator follows immediately:
\\

\begin{corollary}
Suppose the true probability measure $H$ is supported on $\mathscr{C}_{KC}$.
As $n \to \infty$, $\hat{m}_j \overset{p}{\to} m_j$ and $\hat{\sigma}^2_j \overset{p}{\to} \sigma^2_j$;
consequently, $I\{ \theta \in \widehat{\mathscr{C}}_{KC}\} \overset{p}{\to}  I\{ \theta \in \mathscr{C}_{KC} \}$.
\end{corollary}

%In order to selection the complexity of the finite mixture model, we adopt the Bayesian Information Criterion (BIC).  
%As in the main body of this paper, ... we plot $K$ against BIC and clearly see that K=6 and K=7 are reasonable choices.

\if1\blind
{
\section*{Acknowledgments}
This work was partially supported by
NSF BRAIN EAGER award DBI-1451081,
DARPA XDATA contract FA8750-12-2-0303,
DARPA SIMPLEX contract N66001-15-C-4041,
% DARPA GRAPHS contract N66001-14-1-4028,
% JHU HLT COE, %the Johns Hopkins University Human Language Technology Center of Excellence,
% IARPA,
and HHMI Janelia.
% and The Yummy Distinguished Professorship of Watermelon Science.
%The authors would like to thank the Isaac Newton Institute for Mathematical Sciences, Cambridge, UK,
The authors thank the Isaac Newton Institute for Mathematical Sciences, Cambridge, UK,
 for support and hospitality during the programme Theoretical Foundations for Statistical Network Analysis 
 (EPSRC grant no.\ EP/K032208/1)
 where a portion of the work on this paper was undertaken.
The authors thank Keith Levin and Joshua Cape for helpful comments and criticism.
} \fi

%\section*{References}
%
%{\bf citep{WHAT1}}: 
%If the graph is indeed an SBM -- more generally, an RDPG LPM -- then $\mclustase$ provides consistent subsequent inference:\\
%cite STFP, DEF SIMAX, DLS PAMI, Perfect, AA CLT, MT CLT, VN1 AoAS, VN2, VN3, MT AnnStat, ParComp, EB EJS, ...

\bibliographystyle{chicago}
\bibliography{MBstructure}

\begin{thebibliography}{}

\bibitem[\protect\citeauthoryear{Akaike}{Akaike}{1974}]{akaike1974new}
Akaike, H. (1974).
\newblock A new look at the statistical model identification.
\newblock {\em IEEE Transactions on Automatic Control\/}~{\em 19\/}(6),
  716--723.

\bibitem[\protect\citeauthoryear{Athreya, Priebe, Tang, Lyzinski, Marchette,
  and Sussman}{Athreya et~al.}{2016}]{Athreya:2016rc}
Athreya, A., C.~E. Priebe, M.~Tang, V.~Lyzinski, D.~J. Marchette, and D.~L.
  Sussman (2016).
\newblock A limit theorem for scaled eigenvectors of random dot product graphs.
\newblock {\em Sankhya A\/}~{\em 78\/}(1), 1--18.

\bibitem[\protect\citeauthoryear{Bickel and Doksum}{Bickel and
  Doksum}{2007}]{Bickel}
Bickel, P.~J. and K.~A. Doksum (2007).
\newblock {\em {Mathematical Statistics: Basic Ideas and Selected Topics}\/}
  (2nd ed.), Volume~1 of {\em Holden-Day series in probability and statistics}.
\newblock Prentice Hall.

\bibitem[\protect\citeauthoryear{Chatterjee}{Chatterjee}{2015}]{chatterjee2015}
Chatterjee, S. (2015).
\newblock Matrix estimation by universal singular value thresholding.
\newblock {\em The Annals of Statistics\/}~{\em 43\/}(1), 177--214.

\bibitem[\protect\citeauthoryear{Chen, Vogelstein, Lyzinski, and Priebe}{Chen
  et~al.}{2016}]{Chen:2016cb}
Chen, L., J.~T. Vogelstein, V.~Lyzinski, and C.~E. Priebe (2016).
\newblock {A joint graph inference case study: the C. elegans chemical and
  electrical connectomes}.
\newblock {\em Worm\/}~{\em 5\/}(2), e1142041.

\bibitem[\protect\citeauthoryear{Danon, D{\'i}az-Guilera, Duch, and
  Arena}{Danon et~al.}{2005}]{dadiduar05}
Danon, L., A.~D{\'i}az-Guilera, J.~Duch, and A.~Arena (2005).
\newblock Comparing community structure identification.
\newblock {\em Journal of Statistical Mechanics: Theory and Experiment\/}~{\em
  2005\/}(09), P09008.

\bibitem[\protect\citeauthoryear{Diaconis, Goel, and Holmes}{Diaconis
  et~al.}{2008}]{Horseshoes}
Diaconis, P., S.~Goel, and S.~Holmes (2008).
\newblock Horseshoes in multidimensional scaling and local kernel methods.
\newblock {\em The Annals of Applied Statistics\/}~{\em 2\/}(3), 777--807.

\bibitem[\protect\citeauthoryear{Eichler, Li, Kumar, Park, Andrade,
  Schneider-Mizell, Saumweber, Huser, Bonnery, Gerber, Fetter, Truman, Priebe,
  Abbott, Thum, Zlatic, and Cardona}{Eichler et~al.}{2017}]{EichlerSubmitted}
Eichler, K., F.~Li, A.~L. Kumar, Y.~Park, I.~Andrade, C.~Schneider-Mizell,
  T.~Saumweber, A.~Huser, D.~Bonnery, B.~Gerber, R.~D. Fetter, J.~W. Truman,
  C.~E. Priebe, L.~F. Abbott, A.~Thum, M.~Zlatic, and A.~Cardona (2017).
\newblock The complete wiring diagram of a high-order learning and memory
  center, the insect mushroom body.
\newblock {\em Nature\/}, accepted for publication.

\bibitem[\protect\citeauthoryear{Ekman}{Ekman}{1954}]{ekman_54}
Ekman, G. (1954).
\newblock {Dimensions of Color Vision}.
\newblock {\em Journal of Psychology\/}~{\em 38}, 467--474.

\bibitem[\protect\citeauthoryear{Fishkind, Sussman, Tang, Vogelstein, and
  Priebe}{Fishkind et~al.}{2013}]{fishkind2013consistent}
Fishkind, D.~E., D.~L. Sussman, M.~Tang, J.~T. Vogelstein, and C.~E. Priebe
  (2013).
\newblock Consistent adjacency-spectral partitioning for the stochastic block
  model when the model parameters are unknown.
\newblock {\em SIAM Journal on Matrix Analysis and Applications\/}~{\em 34},
  23--39.

\bibitem[\protect\citeauthoryear{Fraley and Raftery}{Fraley and
  Raftery}{2002}]{mclust2012}
Fraley, C. and A.~E. Raftery (2002).
\newblock Model-based clustering, discriminant analysis and density estimation.
\newblock {\em Journal of the American Statistical Association\/}~{\em 97},
  611--631.

\bibitem[\protect\citeauthoryear{Geman and Geman}{Geman and
  Geman}{2016}]{GGPNAS2016}
Geman, D. and S.~Geman (2016).
\newblock Opinion: Science in the age of selfies.
\newblock {\em Proceedings of the National Academy of Sciences\/}~{\em
  113\/}(34), 9384--9387.

\bibitem[\protect\citeauthoryear{Glasser, Coalson, Robinson, Hacker, Harwell,
  Yacoub, Ugurbil, Andersson, Beckmann, Jenkinson, Smith, and
  Van~Essen}{Glasser et~al.}{2016}]{WashU2016}
Glasser, M.~F., T.~S. Coalson, E.~C. Robinson, C.~D. Hacker, J.~Harwell,
  E.~Yacoub, K.~Ugurbil, J.~Andersson, C.~F. Beckmann, M.~Jenkinson, S.~M.
  Smith, and D.~C. Van~Essen (2016).
\newblock A multi-modal parcellation of human cerebral cortex.
\newblock {\em Nature\/}~{\em 536\/}(7615), 171--178.

\bibitem[\protect\citeauthoryear{Hagmann}{Hagmann}{2005}]{3230/THESES}
Hagmann, P. (2005).
\newblock {\em From diffusion {MRI} to brain connectomics}.
\newblock Ph.\ D. thesis, STI, Lausanne.

\bibitem[\protect\citeauthoryear{Hoff, Raftery, and Handcock}{Hoff
  et~al.}{2002}]{Hoff2002}
Hoff, P.~D., A.~E. Raftery, and M.~S. Handcock (2002).
\newblock {Latent space approaches to social network analysis}.
\newblock {\em Journal of the American Statistical Association\/}~{\em
  97\/}(460), 1090--1098.

\bibitem[\protect\citeauthoryear{Holland, Laskey, and Leinhardt}{Holland
  et~al.}{1983}]{holland1983stochastic}
Holland, P.~W., K.~B. Laskey, and S.~Leinhardt (1983).
\newblock Stochastic blockmodels: First steps.
\newblock {\em Social networks\/}~{\em 5\/}(2), 109--137.

\bibitem[\protect\citeauthoryear{Hubert and Arabie}{Hubert and
  Arabie}{1985}]{hubert85}
Hubert, L. and P.~Arabie (1985).
\newblock Comparing partitions.
\newblock {\em Journal of Classification\/}~{\em 2\/}(1), 193--218.

\bibitem[\protect\citeauthoryear{Jaccard}{Jaccard}{1912}]{ja1912}
Jaccard, P. (1912).
\newblock The distribution of the flora in the alpine zone.
\newblock {\em The New Phytologist\/}~{\em 11\/}(2), 37--50.

\bibitem[\protect\citeauthoryear{Jackson}{Jackson}{2004}]{Jackson}
Jackson, J.~E. (2004).
\newblock {\em A User's Guide to Principal Components}.
\newblock John Wiley \& Sons, Inc.

\bibitem[\protect\citeauthoryear{Jain, Duin, and Mao}{Jain
  et~al.}{2000}]{JainDuinMao}
Jain, A.~K., R.~P.~W. Duin, and J.~Mao (2000).
\newblock Statistical pattern recognition: A review.
\newblock {\em IEEE Transactions on Pattern Analysis and Machine
  Intelligence\/}~{\em 22\/}(1), 4--37.

\bibitem[\protect\citeauthoryear{Karrer and Newman}{Karrer and
  Newman}{2011}]{MR2788206}
Karrer, B. and M.~E.~J. Newman (2011).
\newblock Stochastic blockmodels and community structure in networks.
\newblock {\em Physical Review E (3)\/}~{\em 83\/}(1), 016107, 10.

\bibitem[\protect\citeauthoryear{Kiefer and Wolfowitz}{Kiefer and
  Wolfowitz}{1956}]{Kiefer1956}
Kiefer, J. and J.~Wolfowitz (1956).
\newblock Consistency of the maximum likelihood estimator in the presence of
  infinitely many incidental parameters.
\newblock {\em The Annals of Mathematical Statistics\/}~{\em 27}, 886 -- 906.

\bibitem[\protect\citeauthoryear{Ko, Hofer, Pichler, Buchanan, Sjostrom, and
  Mrsic-Flogel}{Ko et~al.}{2011}]{Activity1}
Ko, H., S.~B. Hofer, B.~Pichler, K.~A. Buchanan, P.~J. Sjostrom, and T.~D.
  Mrsic-Flogel (2011).
\newblock Functional specificity of local synaptic connections in neocortical
  networks.
\newblock {\em Nature\/}~{\em 473\/}(7345), 87--91.

\bibitem[\protect\citeauthoryear{Laird}{Laird}{1978}]{Laird1978}
Laird, N. (1978).
\newblock Nonparametric maximum likelihood estimation of a mixing distribution.
\newblock {\em Journal of the American Statistical Association\/}~{\em
  73\/}(364), 805 -- 811.

\bibitem[\protect\citeauthoryear{Lee, Bonin, Reed, Graham, Hood, Glattfelder,
  and Reid}{Lee et~al.}{2016}]{Activity2}
Lee, W.-C.~A., V.~Bonin, M.~Reed, B.~J. Graham, G.~Hood, K.~Glattfelder, and
  R.~C. Reid (2016).
\newblock Anatomy and function of an excitatory network in the visual cortex.
\newblock {\em Nature\/}~{\em 532\/}(7599), 370--374.

\bibitem[\protect\citeauthoryear{Lesperance and Kalbfleisch}{Lesperance and
  Kalbfleisch}{1992}]{Lesperance1992}
Lesperance, M.~L. and J.~D. Kalbfleisch (1992).
\newblock An algorithm for computing the nonparametric mle of a mixing
  distribution.
\newblock {\em Journal of the American Statistical Association\/}~{\em
  87\/}(417), 120 -- 126.

\bibitem[\protect\citeauthoryear{Lindsay}{Lindsay}{1983}]{lindsay1983}
Lindsay, B.~G. (1983).
\newblock The geometry of mixture likelihoods: A general theory.
\newblock {\em The Annals of Statistics\/}~{\em 11\/}(1), 86--94.

\bibitem[\protect\citeauthoryear{Lindsay and Lesperance}{Lindsay and
  Lesperance}{1995}]{Lindsay1995}
Lindsay, B.~G. and M.~L. Lesperance (1995).
\newblock A review of semiparametric mixture models.
\newblock {\em Journal of Statistical Planning and Inference\/}~{\em 47},
  29--39.

\bibitem[\protect\citeauthoryear{Lyzinski, Sussman, Tang, Athreya, and
  Priebe}{Lyzinski et~al.}{2014}]{lyzinski13:_perfec}
Lyzinski, V., D.~L. Sussman, M.~Tang, A.~Athreya, and C.~E. Priebe (2014).
\newblock Perfect clustering for stochastic blockmodel graphs via adjacency
  spectral embedding.
\newblock {\em Electronic Journal of Statistics\/}~{\em 8}, 2905--2922.

\bibitem[\protect\citeauthoryear{Lyzinski, Tang, Athreya, Park, and
  Priebe}{Lyzinski et~al.}{2017}]{Lyzynski2015}
Lyzinski, V., M.~Tang, A.~Athreya, Y.~Park, and C.~E. Priebe (2017).
\newblock Community detection and classification in hierarchical stochastic
  blockmodels.
\newblock {\em IEEE Transactions on Network Science and Engineering\/}~{\em
  4\/}(1), 13--26.

\bibitem[\protect\citeauthoryear{Malliaros and Vazirgiannis}{Malliaros and
  Vazirgiannis}{2013}]{Malliaros201395}
Malliaros, F.~D. and M.~Vazirgiannis (2013).
\newblock Clustering and community detection in directed networks: A survey.
\newblock {\em Physics Reports\/}~{\em 533\/}(4), 95 -- 142.

\bibitem[\protect\citeauthoryear{Meil{\u{a}}}{Meil{\u{a}}}{2007}]{me07}
Meil{\u{a}}, M. (2007).
\newblock Comparing clusterings--an information based distance.
\newblock {\em Journal of Multivariate Analysis\/}, 873--195.

\bibitem[\protect\citeauthoryear{Ohyama, Schneider-Mizell, Fetter, Aleman,
  Franconville, Rivera-Alba, Mensh, Branson, Simpson, Truman, et~al.}{Ohyama
  et~al.}{2015}]{ohyama2015multilevel}
Ohyama, T., C.~M. Schneider-Mizell, R.~D. Fetter, J.~V. Aleman,
  R.~Franconville, M.~Rivera-Alba, B.~D. Mensh, K.~M. Branson, J.~H. Simpson,
  J.~W. Truman, et~al. (2015).
\newblock A multilevel multimodal circuit enhances action selection in
  drosophila.
\newblock {\em Nature\/}~{\em 520\/}(7549), 633--639.

\bibitem[\protect\citeauthoryear{Rissanen}{Rissanen}{1978}]{MDL}
Rissanen, J. (1978).
\newblock Modeling by shortest data description.
\newblock {\em Automatica\/}~{\em 14\/}(5), 465 -- 471.

\bibitem[\protect\citeauthoryear{Schneider-Mizell, Gerhard, Longair, Kazimiers,
  Li, Zwart, Champion, Midgley, Fetter, Saalfeld, et~al.}{Schneider-Mizell
  et~al.}{2016}]{schneider2016quantitative}
Schneider-Mizell, C.~M., S.~Gerhard, M.~Longair, T.~Kazimiers, F.~Li, M.~F.
  Zwart, A.~Champion, F.~M. Midgley, R.~D. Fetter, S.~Saalfeld, et~al. (2016).
\newblock Quantitative neuroanatomy for connectomics in drosophila.
\newblock {\em Elife\/}~{\em 5}, e12059.

\bibitem[\protect\citeauthoryear{Schwarz}{Schwarz}{1978}]{BIC}
Schwarz, G. (1978).
\newblock Estimating the dimension of a model.
\newblock {\em The Annals of Statistics\/}~{\em 6\/}(2), 461--464.

\bibitem[\protect\citeauthoryear{Sporns}{Sporns}{2012}]{Sporns:2012:DHC:2381020}
Sporns, O. (2012).
\newblock {\em Discovering the Human Connectome\/} (1st ed.).
\newblock The MIT Press.

\bibitem[\protect\citeauthoryear{Sporns, Tononi, and Kotter}{Sporns
  et~al.}{2005}]{10.1371/journal.pcbi.0010042}
Sporns, O., G.~Tononi, and R.~Kotter (2005).
\newblock The human connectome: A structural description of the human brain.
\newblock {\em PLoS Computational Biology\/}~{\em 1\/}(4).

\bibitem[\protect\citeauthoryear{Susko, Kalbfleisch, and Chen}{Susko
  et~al.}{1998}]{Susko1998}
Susko, E., J.~D. Kalbfleisch, and J.~Chen (1998).
\newblock Constrained nonparametric maximum-likelihood estimation for mixture
  models.
\newblock {\em Canadian Journal of Statistics\/}~{\em 26\/}(4), 601--617.

\bibitem[\protect\citeauthoryear{Sussman, Tang, Fishkind, and Priebe}{Sussman
  et~al.}{2012}]{STFP2012}
Sussman, D.~L., M.~Tang, D.~E. Fishkind, and C.~E. Priebe (2012).
\newblock A consistent adjacency spectral embedding for stochastic blockmodel
  graphs.
\newblock {\em Journal of the American Statistical Association\/}~{\em
  107\/}(499), 1119--1128.

\bibitem[\protect\citeauthoryear{Sussman, Tang, and Priebe}{Sussman
  et~al.}{2014}]{sussman2012universally}
Sussman, D.~L., M.~Tang, and C.~E. Priebe (2014).
\newblock Consistent latent position estimation and vertex classification for
  random dot product graphs.
\newblock {\em IEEE Transactaions on Pattern Analysis and Machine
  Intelligence\/}~{\em 36}, 48--57.

\bibitem[\protect\citeauthoryear{Tang and Priebe}{Tang and
  Priebe}{2016}]{tang_priebe2016}
Tang, M. and C.~E. Priebe (2016).
\newblock Limit theorems for eigenvectors of the normalized {L}aplacian for
  random graphs.
\newblock {\em arXiv:1607.0123\/}.

\bibitem[\protect\citeauthoryear{Tang, Sussman, and Priebe}{Tang
  et~al.}{2013}]{tang2012universally}
Tang, M., D.~L. Sussman, and C.~E. Priebe (2013).
\newblock Universally consistent vertex classification for latent position
  graphs.
\newblock {\em The Annals of Statistics\/}~{\em 41}, 1406 -- 1430.

\bibitem[\protect\citeauthoryear{Varshney, Chen, Paniagua, Hall, and
  Chklovskii}{Varshney et~al.}{2011}]{10.1371/journal.pcbi.1001066}
Varshney, L.~R., B.~L. Chen, E.~Paniagua, D.~H. Hall, and D.~B. Chklovskii
  (2011).
\newblock Structural properties of the caenorhabditis elegans neuronal network.
\newblock {\em PLoS Computational Biology\/}~{\em 7\/}(2), 1--21.

\bibitem[\protect\citeauthoryear{Vogelstein, Park, Ohyama, Kerr, Truman,
  Priebe, and Zlatic}{Vogelstein et~al.}{2014}]{Vogelstein:2014hn}
Vogelstein, J.~T., Y.~Park, T.~Ohyama, R.~A. Kerr, J.~W. Truman, C.~E. Priebe,
  and M.~Zlatic (2014).
\newblock {Discovery of brainwide neural-behavioral maps via multiscale
  unsupervised structure learning.}
\newblock {\em Science\/}~{\em 344\/}(6182), 386--392.

\bibitem[\protect\citeauthoryear{Wang}{Wang}{2010}]{Wang2010}
Wang, Y. (2010).
\newblock Maximum likelihood computation for fitting semiparametric mixture
  models.
\newblock {\em Statistics and Computing\/}~{\em 20}, 75--86.

\bibitem[\protect\citeauthoryear{Wang and Wong}{Wang and Wong}{1987}]{WangWong}
Wang, Y.~J. and G.~Y. Wong (1987).
\newblock Stochastic blockmodels for directed graphs.
\newblock {\em Journal of the American Statistical Association\/}~{\em 82},
  8--19.

\bibitem[\protect\citeauthoryear{{White}, {Southgate}, {Thomson}, and
  {Brenner}}{{White} et~al.}{1986}]{1986RSPTB.314....1W}
{White}, J.~G., E.~{Southgate}, J.~N. {Thomson}, and S.~{Brenner} (1986).
\newblock {The Structure of the Nervous System of the Nematode Caenorhabditis
  elegans}.
\newblock {\em Philosophical Transactions of the Royal Society of London Series
  B\/}~{\em 314}, 1--340.

\bibitem[\protect\citeauthoryear{Zhu and Ghodsi}{Zhu and
  Ghodsi}{2006}]{Zhu:2006fv}
Zhu, M. and A.~Ghodsi (2006).
\newblock Automatic dimensionality selection from the scree plot via the use of
  profile likelihood.
\newblock {\em Computational Statistics and Data Analysis\/}~{\em 51\/}(2),
  918--930.

\bibitem[\protect\citeauthoryear{Zilles and Amunts}{Zilles and
  Amunts}{2010}]{Brodmann100}
Zilles, K. and K.~Amunts (2010).
\newblock Centenary of {B}rodmann's map -- conception and fate.
\newblock {\em Nature Reviews Neuroscience\/}~{\em 11\/}(2), 139--145.

\end{thebibliography}

\end{document}